\documentclass[review,3p]{elsarticle}

\usepackage{lineno,hyperref}

\usepackage{algorithm}
\usepackage{algorithmic}
\usepackage{eqparbox}

\usepackage[normalem]{ulem}
\useunder{\uline}{\ul}{}
\usepackage[para]{threeparttable}

\usepackage{pgfplots}
\pgfplotsset{compat=1.14}
\usepackage{tikz}
\usepgfplotslibrary{groupplots}
\usetikzlibrary{pgfplots.groupplots,shadows,calc,positioning,shapes.geometric}

\journal{Computers \& Operations Research}

\biboptions{authoryear}

\usepackage{fancyhdr}
\begin{document}

\begin{frontmatter}

\title{\large MineReduce: an approach based on data mining for problem size reduction\tnoteref{t1}}
\tnotetext[t1]{This manuscript has been accepted for publication on May 5, 2020. Publication DOI: \href{https://doi.org/10.1016/j.cor.2020.104995}{10.1016/j.cor.2020.104995}.}


\author[uff]{Marcelo Rodrigues de Holanda Maia\corref{correspondingauthor}}
\cortext[correspondingauthor]{Corresponding author}
\ead{mmaia@ic.uff.br}

\author[uff]{Alexandre Plastino}
\ead{plastino@ic.uff.br}

\author[ufop]{Puca Huachi Vaz Penna}
\ead{puca@ufop.edu.br}

\address[uff]{Instituto de Computa\c{c}\~{a}o, Universidade Federal Fluminense, Avenida General Milton Tavares de Souza s/n, Niter\'{o}i, RJ, Brazil, 24210-346}
\address[ufop]{Instituto de Ci\^{e}ncias Exatas e Biol\'{o}gicas, Universidade Federal de Ouro Preto, Campus Universit\'{a}rio Morro do Cruzeiro, Ouro Preto, MG, Brazil, 35400-000}

\begin{abstract}
Hybrid variations of metaheuristics that include data mining strategies have been utilized to solve a variety of combinatorial optimization problems, with superior and encouraging results. Previous hybrid strategies applied mined patterns to guide the construction of initial solutions, leading to more effective exploration of the solution space. Solving a combinatorial optimization problem is usually a hard task because its solution space grows exponentially with its size. Therefore, problem size reduction is also a useful strategy in this context, especially in the case of large-scale problems. In this paper, we build upon these ideas by presenting an approach named MineReduce, which uses mined patterns to perform problem size reduction. We present an application of MineReduce to improve a heuristic for the heterogeneous fleet vehicle routing problem. The results obtained in computational experiments show that this proposed heuristic demonstrates superior performance compared to the original heuristic and other state-of-the-art heuristics, achieving better solution costs with shorter run times.
\end{abstract}

\begin{keyword}
Problem size reduction\sep Hybrid metaheuristics\sep Data mining\sep Combinatorial optimization\sep Vehicle routing
\end{keyword}

\end{frontmatter}


\section{Introduction}
\label{intro}

Hybrid metaheuristics have been proposed and successfully applied to combinatorial optimization problems (COPs) in several areas, allowing near-optimal solutions to be found within an acceptable computational time frame. One successful example is the hybridization of the Greedy Randomized Adaptive Search Procedure (GRASP) metaheuristic \citep{resende_ribeiro2016} with data mining techniques \citep{martinsetal2018b}.

Previous work that explored data science in combinatorial optimization produced highly significant results for several problems~\citep{ribeiroetal2006,santosetal2008,plastinoetal2011,plastinoetal2014,barbalhoetal2013,guerineetal2016,maiaetal2018,martinsetal2018a}. Data-mining-hybridized heuristics were able to find solutions of higher quality while spending less computational time when compared to their non-hybridized counterparts and other state-of-the-art heuristics. They applied patterns (found by data mining procedures) to guide the construction of initial solutions.

The difficulty in solving a COP is due to the fact that its solution space grows exponentially with its size. Therefore, problem size reduction (PSR) -- which consists of reducing the size of a problem instance, then solving the reduced instance and expanding it back -- has been found to be a very useful strategy in this context, especially in the case of large-scale problems, since it can significantly reduce the search space of a COP~\citep{gavish_srikanth1986,fischer_merz2007,delgadilloetal2016}.

Building upon these ideas, in this paper, we present an approach named MineReduce, which uses mined patterns to perform problem size reduction. Previous methods that also incorporate data mining techniques into metaheuristics, in general, use the mined patterns as a starting point for the construction of initial solutions \citep{martinsetal2018b}. MineReduce, on the other hand, performs a PSR procedure by removing or merging elements that are in a pattern, finding a solution to the reduced instance, and expanding the solution found, which will be used as the starting point for a local search on the original solution space.

In order to validate the MineReduce approach, we have applied it to extend a previous and state-of-the-art heuristic for the heterogeneous fleet vehicle routing problem (HFVRP), obtaining significantly better results in terms of both solution quality and computational time.

So, the main contribution of this work is twofold: an approach based on the idea of using mined patterns to perform PSR, and a state-of-the-art heuristic for solving the HFVRP based on this approach. The outcomes of computational experiments carried out on a large quantity of HFVRP instances from extensively used collections affirm the effectiveness of the proposed heuristic. The results show that the proposed heuristic based on MineReduce demonstrates superior performance when compared with a state-of-the-art heuristic proposed by \citet{pennaetal2013a}, which served as a baseline for the development of our proposal, and also with a recent hybrid data mining approach proposed for the problem~\citep{maiaetal2018}, achieving better solution costs with shorter run times. Furthermore, new best solutions to 22 instances were found.

Additionally, we have compared the results obtained by our proposed heuristic to those reported by \citet{kochetov_khmelev2015} and \citet{pennaetal2019} for their state-of-the-art algorithms. MineReduce proved to be very competitive in this comparison, especially for large instances.

The remainder of this article is organized as follows. Section~\ref{sec:relatedWork} presents a brief review of related work. Section~\ref{sec:novelApproach} introduces the MineReduce approach. Section~\ref{sec:novelHeuristics} describes the application of the MineReduce approach to the HFVRP. In Section~\ref{sec:results}, the outcomes from experiments using the MineReduce-based heuristic are analyzed and compared with those obtained using previous heuristics. Finally, Section~\ref{sec:conclusion} provides conclusions and directions for future work.

\section{Related work} \label{sec:relatedWork}

This section presents a brief review of related work regarding hybrid data mining heuristics (Section~\ref{subsec:rwDMHeuristics}) and PSR in the context of combinatorial optimization (Section~\ref{subsec:rwPSReduction}).

\subsection{Hybrid data mining heuristics} \label{subsec:rwDMHeuristics}

One successful example of hybrid data mining heuristic is the proposal of a hybrid version of the GRASP metaheuristic which incorporates a data mining module \citep{ribeiroetal2006}, called Data Mining GRASP (DM-GRASP). The basic idea of this hybrid metaheuristic is that patterns found in good solutions can be used to guide the search, leading to more effective exploration of the solution space. In this hybrid version, after the execution of half of the GRASP iterations, a data mining procedure is applied to extract patterns from an elite set composed of the best solutions found up to that point. These patterns represent features found in the elite set of solutions and are used to guide the search in the second half of the iterations. It has been successfully applied to the set packing \citep{ribeiroetal2006}, maximum diversity \citep{santosetal2005}, server replication for reliable multicasting \citep{santosetal2006}, \textit{p}-median \citep{plastinoetal2011,martinsetal2018a}, 2-path network design \citep{barbalhoetal2013} and one-commodity pickup-and-delivery traveling salesman \citep{guerineetal2016} problems. A survey of this approach has been presented by \citet{santosetal2008}.

Subsequently, a new version of the DM-GRASP metaheuristic, called Multi-Data Mining GRASP (MDM-GRASP), was proposed \citep{plastinoetal2014}. The main idea underlying this version is to run the data mining procedure multiple times in an adaptive mode. MDM-GRASP has also been applied to the \textit{p}-median \citep{plastinoetal2011}, 2-path network design \citep{barbalhoetal2013}, server replication for reliable multicasting \citep{plastinoetal2014} and one-com\-mod\-i\-ty pickup-and-delivery traveling salesman \citep{guerineetal2016} problems, and it has a\-chieved better results than those obtained by DM-GRASP, not only in terms of solution quality but also concerning computational time.

Recently, the data mining techniques used in DM-GRASP and MDM-GRASP have been incorporated into a multi-start ILS (MS-ILS) heuristic for the HFVRP, resulting in the state-of-the-art hybrid heuristics DM-MS-ILS and MDM-MS-ILS \citep{maiaetal2018}.

A survey of heuristics that incorporate data mining procedures was presented by \citet{martinsetal2018b}.

\subsection{Problem size reduction} \label{subsec:rwPSReduction}

Any approach for solving a COP relies on some form of search on its solution space, which is the domain of the function to be optimized. Solving a COP is usually a hard task because its solution space grows exponentially with its size. Therefore, PSR is beneficial in this context, especially in the case of large-scale COPs~\citep{gavish_pirkul1985a, gavish_pirkul1985b, gavish_srikanth1986}.

The size of a COP is defined by its decision variables, i.e., the dimensions of its solution space. Hence, PSR techniques have the aim of reducing the number of decision variables of the problem.

The general process of PSR techniques consists of: (i) transforming a problem $P$ into a modified problem $P'$ such that the number of decision variables of $P'$ is smaller than the number of decision variables of $P$, (ii) solving $P'$, and (iii) transforming the solution to $P'$ into a solution to $P$.

The application of this procedural framework is exemplified by the Reduce-Optimize-Expand (ROE) method \citep{montieletal2013}. In the first step (reduce), the ROE method reduces the size of problem instances just before the application of a solving algorithm. Once the reduction is achieved, the next step (optimize) is the application of a combinatorial optimization method, exact or heuristic, to obtain an optimal or suboptimal solution. Then it executes the last step (expand) to obtain the final result.

Size reduction is naturally the most crucial procedure in the PSR process since better decisions in this step will produce solutions of better quality in the subsequent stages \citep{montieletal2013,montiel_delgadillo2015,delgadilloetal2016}. Strategies to make these decisions are highly dependent on the structure and features of the problem, so they vary significantly among distinct classes of COPs.

One general form of accomplishing size reduction is fixing values of decision variables, which can be regarded as fixing elements either in or out of the solution. This kind of reduction is common for many classes of COPs, like routing problems, which include the HFVRP. For example, in binary formulations of classical routing problems -- such as the traveling salesman problem (TSP) or the vehicle routing problem (VRP) -- the decision variables refer to edges in the problem instance graph. The value set to a variable indicates whether the corresponding edge is in (one) or out of (zero) the solution.

Analyzing routing problems as graph-based models, which is typical for this class of problems, approaches to reduce the size of an instance can be either vertex-based or edge-based, as stated by \citet{fischer_merz2007} for the TSP. In a vertex-based approach, subsets of vertices are merged into one single vertex or cluster-vertex. In an edge-based approach, a sequence of edges is merged into one single edge or fixed-path-edge. In both cases, what is implicitly done is also fixing the values of decision variables.

Studies found in the literature in which PSR has been successfully applied to routing problems include those by \citet{gavish_srikanth1986}, \citet{min1991}, \citet{barnhartetal2002}, \citet{fischer_merz2007}, \citet{ramos_oliveira2011}, \citet{lin2011}, \citet{montieletal2013}, \citet{montiel_delgadillo2015} and \citet{delgadilloetal2016}.

Work on the multilevel paradigm, which involves the recursive application of the PSR steps, has provided evidence that it can aid metaheuristics to find better solutions faster for some COPs~\citep{walshaw2008}. In a multilevel optimization method, the original problem instance is recursively coarsened (reduced), creating a multilevel hierarchy of reductions. An initial solution is found (at the coarsest level) and then, at each level in reverse order, iteratively expanded to a solution to the parent level's instance and refined, usually with a local search algorithm.

\citet{chen2015} proposed a fix-and-optimize approach for mixed integer programming problems. It decomposes the original problem by fixing values of binary variables based on their interrelatedness. That approach has also been integrated into a variable neighborhood search framework, achieving excellent results for lot sizing~\citep{chen2015,lietal2017} and transportation~\citep{toschietal2018} problems.

\citet{blumetal2016} introduced a hybrid metaheuristic called Construct, Merge, Solve \& Adapt (CMSA). It is based on the idea of generating solutions, reducing the original instance by merging the components in the generated solutions (in a way such that a solution to the original instance can be derived from a solution to a reduced instance), solving the reduced instances to optimality, and adapting these reduced instances based on an ageing mechanism. The CMSA metaheuristic has achieved good results for some COPs, like the minimum common string partition \citep{blumetal2016,blum_raidl2016}, the minimum covering arborescence \citep{blumetal2016}, and the repetition-free longest common subsequence \citep{blum_blesa2016}.

\citet{kennyetal2019} presented another hybrid metaheuristic, based on a strategy known as ``merge search", which has some similarities with CMSA. The main difference between them is that, while CMSA generates solutions from scratch, the merge search algorithm starts with a single initial seed solution and uses local search to generate a population of neighbouring solutions to the initial seed solution. Then it goes through an iterative process of generating a population, merging, and solving the reduced instance using an exact method. The merging procedure is based on the intersections between all of the solutions in the population. The solution to a reduced instance becomes the initial seed solution to the next generation. As each iteration produces a new set of solutions, there is no need for an ageing mechanism to regulate the population. This metaheuristic has achieved excellent results for the constrained pit problem.

\section{The MineReduce approach} \label{sec:novelApproach}

Several multi-start heuristics, such as the GRASP-based ones, perform a sequence of independent iterations, composed of a generation phase and a local search phase. The generation phase builds an initial solution and, in general, consists of the application of more straightforward methods, usually based on a combination of greedy and random strategies. Most of the computational effort is employed in the local search phase, which improves the initial solution.

Previous approaches that incorporate data mining into these heuristics eliminate the independence of their iterations by introducing a memory mechanism into them. At some point, iterations begin to benefit from knowledge accumulated in previous iterations. This knowledge -- patterns mined from an elite set of solutions -- is used to generate better initial solutions.

The data mining procedure in these approaches relies on the FPmax* algorithm~\citep{grahne_zhu2003}, which mines maximal frequent itemsets. An itemset is considered to be frequent if it achieves a given minimum support, i.e., if it is present in at least a given minimum number of the elite set solutions. Hence, mined patterns are composed of items that frequently appear together in the sub-optimal solutions of the elite set. Intuitively, it is assumed these items should likely be part of the best solutions to the problem. Thus, they are included in initial solutions.

With the MineReduce approach, we build upon these ideas. Since we assume the mined patterns should likely be part of the best solutions to a problem instance, then they could be well-suited for reducing the size of that instance. The items in a pattern could, for example, be fixed in the solution, which in turn would be equivalent to fixing values of decision variables related to those items. Another possibility is that the items in a pattern could be merged in a condensed representation, also producing a reduced-size instance.

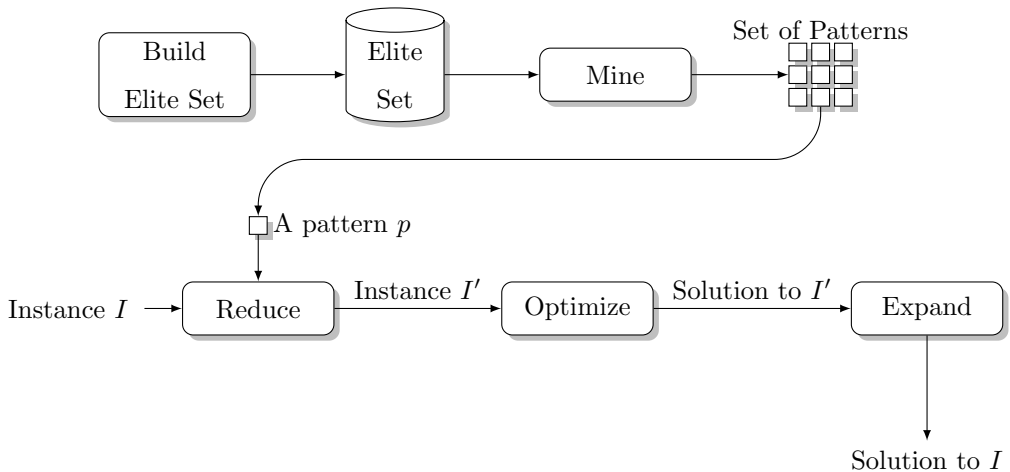
\begin{figure}[!ht]

  \centering

\tikzstyle{block} = [rectangle, draw, fill=white, 
    text width=5em, text centered, rounded corners, node distance=2cm, minimum height=2em,drop shadow]
\tikzstyle{label} = [rectangle, 
    text width=7em, text centered, node distance=2cm]
\tikzstyle{pattern} = [rectangle, draw, fill=white, 
    node distance=2cm,drop shadow]
\tikzstyle{database} = [cylinder,
      shape border rotate=90,
      aspect=0.25,
      draw,
      fill=white, text width=3em, text centered, node distance=2cm, drop shadow]
\tikzstyle{line} = [draw, -latex]

\begin{tikzpicture}[node distance = 3cm, auto]
    \node [block]                         (buildES) {Build Elite Set};
    \node [database, right of = buildES, xshift=0.9cm]     (ES)    {Elite Set};
    \node [block, right of = ES, xshift=0.9cm]     (mine)    {Mine};
    \node [pattern, right of = mine, xshift=0.4cm]     (p10)    {};
    \node [pattern, above of = p10, yshift=-1.7cm]     (p00)    {};
    \node [pattern, below of = p10, yshift=1.7cm]     (p20)    {};
    \node [pattern, right of = p00, xshift=-1.7cm]     (p01)    {};
    \node [pattern, right of = p10, xshift=-1.7cm]     (p11)    {};
    \node [pattern, right of = p20, xshift=-1.7cm]     (p21)    {};
    \node [pattern, right of = p01, xshift=-1.7cm]     (p02)    {};
    \node [pattern, right of = p11, xshift=-1.7cm]     (p12)    {};
    \node [pattern, right of = p21, xshift=-1.7cm]     (p22)    {};
    \node [label, above of = p01, yshift=-1.7cm]     (SP)    {Set of Patterns};
    \node [pattern, below of = ES, xshift=-1.8cm]     (p)    {};
    \node [label, right of = p, xshift=-0.9cm]     (pp)    {A pattern $p$};
    \node [block, below of = p, yshift=0.9cm]     (reduce)    {Reduce};
    \node [label, left of = reduce, xshift=-0.5cm, text width=5em]     (i)    {Instance $I$};
    \node [block, right of = reduce, xshift=2.2cm]     (optimize)    {Optimize};
    \node [block, right of = optimize, xshift=2.6cm]     (expand)    {Expand};
    \node [label, below of = expand]     (sol)    {Solution to $I$};
    \path [line] (buildES) -- (ES);
    \path [line] (ES) -- (mine);
    \path [line] (mine) -- (p10);
    \path [line,rounded corners=4ex] 
          (p21.south) -- ++(0,-0.7cm) -| (p.north);
    \path [line] (p) -- (reduce);
    \path [line] (i) -- (reduce);
    \path [line] (reduce) -- (optimize) node[above, xshift=-2.1cm] {Instance $I'$};
    \path [line] (optimize) -- (expand) node[above, xshift=-2.3cm] {Solution to $I'$};
    \path [line] (expand) -- (sol);
\end{tikzpicture}

\caption{Framework of the MineReduce approach}
\label{fig:framework}

\end{figure}

Fig.~\ref{fig:framework} presents the general framework of the MineReduce approach. Its first steps are to \emph{build an elite set} of solutions and to \emph{mine} patterns from this set. These steps are supposed to be done like in the previous approaches \citep{martinsetal2018b}, i.e., the best solutions found are stored in the elite set until it becomes stable (unaltered for a given period), which triggers the data mining procedure.

The subsequent steps compose a full PSR process, which is intended to replace the heuristic's initial solution generation procedure. The \emph{Reduce} step uses a pattern $p$ to transform a problem instance $I$ into a reduced-size instance $I'$. The \emph{Optimize} step is accomplished through the application of the heuristic's original optimization procedures to~$I'$. In the \emph{Expand} step, the solution to $I'$ is transformed into a solution to $I$, which concludes the MineReduce-based generation of an initial solution.

This modified constructive method is what makes MineReduce different from the previous approaches that also incorporate data mining techniques into metaheuristics. Previous methods derived from the DM-GRASP and MDM-GRASP metaheuristics -- e.g., the hybrid heuristics for the HFVRP proposed in \citep{maiaetal2018} -- use the mined patterns only as a starting point for the construction of initial solutions. MineReduce, on the other hand, performs a procedure that reduces the original instance by deleting or merging elements that are in a pattern, finds a solution to the reduced instance and expands the solution found, which will be used as the starting point for a local search on the original solution space.

In this sense, the MineReduce approach shares some aspects with the multilevel paradigm~\citep{walshaw2008}. Each iteration after the first call to the data mining procedure reduces the problem instance to a ``coarser" level, finds a solution to the reduced instance, expands the solution (back to the original instance's level) and refines it with a local search. In this case, the number of levels is constant (two), but the main difference resides in the reduction strategy. In the multilevel paradigm, this is often done through adaptation of construction heuristics, whereas the MineReduce approach relies on the use of patterns extracted from an elite set of solutions through data mining techniques.

Regarding the reduction strategy, the MineReduce approach is closer to merge search algorithms~\citep{kennyetal2019}, which reduce instances by merging variables based on the intersections observed in a set of solutions. MineReduce's reduction strategy also relies on similarities observed in a set of solutions. However, they differ significantly in how they build such a set of solutions and in how similarities between its solutions are characterized and identified.

\section{Application of the MineReduce approach to the HFVRP} \label{sec:novelHeuristics}

This section presents the application of the MineReduce approach to extend the MS-ILS heuristic for the HFVRP proposed by~\citet{pennaetal2013a}.

\subsection{The HFVRP} \label{sec:problem}

The HFVRP is described as follows. Let $G=(V,A)$ be a directed graph, where $V=\{0,1,...,n\}$ is a set composed of $n+1$ vertices and $A=\{(i,j):i,j \in V, i \neq j \}$ is the set of arcs. Vertex $0$ is the depot, where the vehicle fleet is located, whereas the set $V'=V\backslash\{0\}$ consists of the remaining vertices representing the $n$ customers. Each customer $i \in V'$ is associated with a non-negative demand $q_i$. The fleet consists of $m$ distinct vehicle types, which compose a set $M=\{1,...,m\}$. For each vehicle type $u \in M$, there are $m_u$ vehicles available, each with a capacity $Q_u$ and a fixed cost $f_u$. Finally, for each combination of an arc $(i,j) \in A$ and a vehicle type $u \in M$, there is an associated cost $c^u_{ij}=d_{ij}r_u$, where $d_{ij}$ is the distance between vertices $i$ and $j$ and $r_u$ is the dependent (variable) cost per unit distance associated with the vehicle type $u$.

A route is defined by a pair $(R,u)$, where $R=(i_1,i_2,...,i_{|R|})$, $i_1=i_{|R|}=0$, and $\{i_2,...,i_{|R|-1}\} \subseteq V'$; that is, each route is a circuit in $G$ that starts and ends at the depot and is assigned to a vehicle of type $u \in M$. A route $(R,u)$ is feasible if the sum of the demands of all customers on $R$ does not exceed the capacity $Q_u$ of the vehicle assigned to it. The cost of a route is the sum of the fixed cost of the assigned vehicle and the dependent costs associated with each of the traversed arcs in combination with the vehicle type. Thus, the aim is to discover feasible routes such that each customer is visited precisely once, the total quantity of routes assigned to vehicles of each type $u \in M$ does not exceed $m_u$, and the sum of all route costs is minimized.

HFVRP instances are typically categorized with respect to certain criteria. Two primary classes are related to the limitations on the fleet. The problem that characterizes the first class, known as the Fleet Size and Mix (FSM) problem \citep{goldenetal1984}, can be considered as a particular case of the above definition in which $m_u=\infty, \forall u \in M$. Therefore, this problem consists of identifying the best composition of the fleet as well as its best routing scheme. For the second category of instances, in which the fleet is limited, the corresponding problem is known as the Heterogeneous Fixed Fleet VRP (HFFVRP) \citep{taillard1999} and consists of optimizing the routing for an available fixed fleet.

The FSM and HFFVRP classes may be further subdivided with respect to the types of vehicle costs considered. The possibilities are as follows: both fixed and dependent costs (the general case), fixed costs only (a particular case in which $r_u=1, \forall u \in M$), or dependent costs only (a particular case in which $f_u=0, \forall u \in M$). Five subclasses of FSM and HFFVRP instances with respect to this criterion are differentiated in the literature: (1) the FSM problem with both fixed and dependent costs, denoted by FSM-FD \citep{ferland_michelon1988}; (2) the FSM problem with fixed costs only, denoted by FSM-F \citep{goldenetal1984}; (3)~the FSM problem with dependent costs only, denoted by FSM-D \citep{taillard1999}; (4) the HFFVRP with both fixed and dependent costs, denoted by HFFVRP-FD \citep{lietal2007}; and (5) the HFFVRP with dependent costs only, denoted by HFFVRP-D \citep{taillard1999}. To the best of our knowledge, the HFFVRP with fixed costs only has never been addressed.

An overview of the strategies in the literature for solving the HFVRP can be found in a survey by \citet{baldaccietal2008}. A literature survey of this problem that focuses on industrial aspects of the FSM problem has been presented by \citet{hoffetal2010}. A more recent literature survey on the HFVRP has been presented by \citet{kocetal2016}. The latter provides a computational comparison among state-of-the-art algorithms. It indicates that the best performances for the FSM problem had been achieved by the algorithms proposed by \citet{choi_tcha2007}, \citet{pennaetal2013a} and \citet{vidaletal2014}, whereas the best performances for the HFFVRP had been achieved by the algorithms proposed by \citet{lietal2007}, \citet{liu2013}, \citet{pennaetal2013a} and \citet{subramanianetal2012}. After publication of the survey by \citet{kocetal2016}, other heuristic strategies presented competitive results \citep{maiaetal2018,pennaetal2019} and an exact method found new optimal solutions for some instances~\citep{pessoaetal2018}.

Multi-start heuristics based on the Iterated Local Search (ILS) metaheuristic, in particular, have presented competitive results when applied to vehicle routing problems \citep{subramanianetal2012,pennaetal2013a,pennaetal2013b,pennaetal2019}. Recently, hybrid heuristics for the HFVRP based on the incorporation of data mining strategies into the state-of-the-art MS-ILS heuristic of \citet{pennaetal2013a} were proposed and demonstrated superior performance compared with the original heuristic, achieving better solution costs with shorter run times~\citep{maiaetal2018}. Therefore, the MS-ILS heuristic, described in the next section, was chosen as the basis for the application of the MineReduce approach. We compared the heuristic proposed in this work to the MS-ILS heuristic of \citet{pennaetal2013a} and to its hybrid version called MDM-MS-ILS, presented in~\citep{maiaetal2018}.

\subsection{MS-ILS heuristic} \label{sec:previousApproach}

In this section, we describe -- in a high level of abstraction (we suppress details that are irrelevant for the context of this work) -- the MS-ILS (Multi-Start Iterated Local Search) heuristic proposed by \citet{pennaetal2013a} for solving the HFVRP, which served as a basis for the MineReduce heuristic proposed in this work. The steps of the MS-ILS are presented in Algorithm~\ref{alg:MSILS}.

\begin{algorithm}[!ht]
	\caption{MS-ILS$(MaxIter)$}
	\label{alg:MSILS}
	\begin{algorithmic}[1]
	\STATE $f(s^*) \leftarrow \infty$
	\FOR{$i \leftarrow 1$ \textbf{to} $MaxIter$}
		\STATE $s \leftarrow$ GenerateInitialSolution$()$
		\STATE $s' \leftarrow$ ILS$(s)$
		\IF{$f(s') < f(s^*)$} \STATE $s^* \leftarrow s'$ \ENDIF
	\ENDFOR
	\STATE \textbf{return} $s^*$
	\end{algorithmic}
\end{algorithm}

The multi-start heuristic is run for $MaxIter$ iterations (lines 2--8). In each iteration, a constructive procedure (GenerateInitialSolution$()$) applies a mix of simple greedy and randomized strategies for producing an initial solution (line 3). In the local search phase, the ILS heuristic is used to enhance the generated initial solution (line 4). If the solution $s'$ returned by the ILS heuristic represents an improvement in cost, as given by the function $f$, then the best overall solution $s^*$ is updated (lines 5--7). After the execution of the $MaxIter$ multi-start iterations, the best solution found is returned (line 9).

\subsection{MineReduce-MS-ILS}

The high-level structure of a general MineReduce-based multi-start metaheuristic (MineReduce-MS) is presented in Algorithm \ref{alg:MRMDMMSILS}. It uses the same strategies from the MDM-GRASP metaheuristic \citep{plastinoetal2014} for building the elite set and triggering the data mining procedure. Its more distinctive aspect is the use of a MineReduce-based initial solution generation procedure (line 12).

\begin{algorithm}[!ht]
	\caption{\textsc{MineReduce-MS$(MaxIter,d,MaxP,MinSup,\delta)$}}
	\label{alg:MRMDMMSILS}
	\begin{algorithmic}[1]
	\STATE $f(s^*) \leftarrow \infty$
	\STATE $E \leftarrow$ {\O}
	\STATE $P \leftarrow$ {\O}
	\FOR{$i \leftarrow 1$ \textbf{to} $MaxIter$}
		\IF{Stable$(E,\delta)$}
		    \STATE $P \leftarrow$ Mine$(E,MaxP,MinSup)$
		\ENDIF
		\IF{$P=$ {\O}} \STATE $s \leftarrow$ GenerateInitialSolution$()$
		\ELSE
			\STATE $p \leftarrow$ NextPattern$(P)$
			\STATE $s \leftarrow$ MineReduceGeneration$(p)$ \COMMENT{MineReduce-based generation phase}
		\ENDIF
		\STATE $s' \leftarrow$ LocalSearch$(s)$
		\STATE UpdateEliteSet$(E,s',d)$
		\IF{$f(s') < f(s^*)$} \STATE $s^* \leftarrow s'$ \ENDIF
	\ENDFOR
	\STATE \textbf{return} $s^*$
	\end{algorithmic}
\end{algorithm}

Like in several typical multi-start metaheuristics, each iteration in this algorithm is composed of a generation phase (lines 8--13) and a local search phase (line 14). However, some differences can be observed. Whenever the elite set~$E$ becomes stable the data mining procedure is triggered -- if the data mining procedure has not been called yet, $E$ is stable if it has not been modified in the last $\delta$ iterations; otherwise, $E$ is stable if it has been modified after the last call to the data mining procedure but has not been modified in the last $\delta$ iterations. The data mining procedure returns a set $P$ composed of the $MaxP$ largest patterns found with the minimum support $MinSup$ (lines 5--7), which are arranged in decreasing order by size, forming a circular list. In the early iterations, when data mining has not yet been carried out and the pattern set $P$ is therefore still empty, the initial solutions are generated through a traditional constructive procedure, based on a combination of greedy and random strategies (line 9). Once data mining has been performed, each initial solution is then generated through the MineReduce-based procedure (MineReduceGeneration$(p)$) using a pattern $p$ picked from the current pattern set $P$ following the sequence of the circular list (lines 11--12). After the local search phase (line 14), the elite set is updated (line~15), as is the best solution in case of improvement (lines 16--18).

The pseudo-code of the MineReduce-based constructive method, which implements the PSR steps defined in Fig. \ref{fig:framework}, is presented in Algorithm \ref{alg:MRGeneration}. It is worth observing that the processed instance $I$ is regarded as an implicit input.

\begin{algorithm}[!ht]
	\caption{\textsc{MineReduceGeneration$(p)$}}
	\label{alg:MRGeneration}
	\begin{algorithmic}[1]
	\STATE $\mu \leftarrow$ ReduceInstance$(p)$ \COMMENT{\emph{Reduce} step -- $I$ is reduced into $I'$}
	\STATE $s' \leftarrow$ GenerateInitialSolution$()$ \COMMENT{\emph{Optimize} step (lines 2--3) -- over $I'$}
	\STATE $s' \leftarrow$ LocalSearch$(s')$
	\STATE $s \leftarrow$ ExpandSolution$(s',\mu)$ \COMMENT{\emph{Expand} step -- generates a solution to $I$}
	\STATE \textbf{return} $s$
	\end{algorithmic}
\end{algorithm}

In the \emph{Reduce} step, the instance is reduced based on the provided pattern $p$, returning a map $\mu$ that associates elements of the reduced instance to their corresponding elements in the original instance~(line~1). The \emph{Optimize} step consists of the application of one complete iteration of the multi-start metaheuristic (generation and local search) to the reduced instance. An initial solution to the reduced instance is generated~(line 2), and then a local search starting from it is performed (line 3). Finally, in the \emph{Expand} step, the solution found in the local search is expanded based on the map $\mu$~(line~4) and returned (line 5).

The proposed MineReduce-MS-ILS heuristic is the application of the MineReduce approach on the MS-ILS proposed by \citet{pennaetal2013a}. It is, therefore, an implementation of Algorithm \ref{alg:MRMDMMSILS} where the local search (line~14) is performed by the ILS procedure from Algorithm~\ref{alg:MSILS}.

In this proposal, solutions in the elite set are represented as sets of arcs instead of sequences of vertices. This representation allows the application of the data mining procedure to extract maximal frequent itemsets. As described in Section~\ref{sec:problem}, each route in the HFVRP is represented by a pair $(R,u)$, where $R=(i_1,i_2,...,i_{|R|})$ is a list of vertices, ordered according to the defined visiting sequence, and $u$ is the type of vehicle assigned to the route. In the adopted alternative representation, for each route $(R,u)$, the list $R$ is decomposed into a set of arcs $\{(i_1,i_2),(i_2, i_3),...,(i_{|R|-1},i_{|R|})\}$. Then, the vehicle type $u$ is assigned to each arc in the set, resulting in a set in which each element is a pair composed of an arc $(i_r,i_{r+1})$, $r=1,2,...,|R|-1$, in $R$ and the vehicle type $u$, with the form $\{((i_1,i_2),u),((i_2, i_3),u),...,((i_{|R|-1},i_{|R|}),u)\}$. An arc must be present and associated with the same vehicle type in a minimum number of solutions in the elite set to belong to a pattern. Consequently, the patterns mined are formed of route segments, each one with a vehicle type assigned.

This pattern mining scheme for the HFVRP has been adopted in hybrid versions of the MS-ILS heuristic proposed in~\citep{maiaetal2018}. However, these previous MS-ILS extensions used mined patterns only as a starting point for the construction of initial solutions. Specifically, the constructive procedure used the route segments of a mined pattern, with their respective vehicle types assigned, as the initial routes of the solution, and afterwards completed the solution by inserting the remaining customers.

In MineReduce-MS-ILS, the \emph{Reduce} step of MineReduce relies on the use of patterns mined from the elite set to perform a vertex-based PSR procedure by merging customer vertices that appear consecutively (in the same route segment) in a pattern into one customer cluster vertex.

In a binary formulation of the HFVRP, decision variables are defined as: $x_{ij}^k = 1$ if a vehicle of type $k$ travels from customer $i$ to customer $j$; and $x_{ij}^k = 0$ otherwise. Therefore, differently from the classic VRP, in this case, a vertex-based PSR procedure is not equivalent to fixing values of decision variables. Instead, the original set of decision variables is replaced by a smaller one (since the set of customers is reduced), and the values of the variables in the original set are defined by the values of the corresponding variables in the reduced set.

For using this strategy, it is necessary to extend the HFVRP model presented in Section~\ref{sec:problem}. Since it must be possible to represent a route segment as a customer vertex, each customer $i \in V$ must have an associated $l_i$ value corresponding to the length of the underlying route segment. This value is used to calculate the cost $c^u_i = l_{i}r_u$, which represents the variable cost for a vehicle of type $u$ to traverse the route segment represented by $i$, for each combination of a customer $i \in V$ and a vehicle type $u \in M$. Customer vertices of regular (non-reduced) instances, as well as customer vertices of reduced instances that are not customer cluster vertices, have length $0$. The cost of a route becomes, in this extended model, the sum of the fixed cost of the vehicle associated with the route and the variable costs associated (i) with the combination between the vehicle type and each of the traversed arcs, and (ii) with the combination of the type of vehicle and each of the visited customer vertices.

Let $G = (V, A)$ be a directed graph associated with an HFVRP instance and $p$ a pattern consisting of a set of route segments in that instance, each segment defined by a pair $(R', u)$, where $R'= (i_1, i_2, ..., i_{|R'|})$. Let $G^* = (V^*, A^*) $ be a directed graph associated with a reduced version of the instance associated with $G$ based on $p$. Such a reduced version can be obtained as follows. Initially, $G^*$ is defined as a copy of $G$. For each route segment $(R', u) \in p$, each of the customers in $R'$ is removed from $G^*$ -- that is, the vertex corresponding to the customer is removed from $V^*$ and the arcs that connect that vertex to the others are removed from $A^*$. Also, a customer cluster vertex $i_{R'}$ corresponding to the route segment is added to $V^*$ and arcs connecting $i_{R'}$ to the other vertices in $V^*$ are added to $A^*$. The demand for $i_{R'}$ is given by $q_{i_{R'}}=\sum\limits_{r'=1}^{|R'|}{q_{i_{r'}}}$, that is, the sum of customer demands in $R'$. The distance from each vertex $i^* \in V^*$ to $i_{R'}$ is given by $d_{i^*i_{R'}}=d_{i^*i_1}$, that is, the distance from $i^*$ to $i_1$ (the first customer in $R'$). The distance from $i_{R'}$ to each vertex $i^* \in V^*$ is given by $d_{i_{R'}i^*}=d_{i_{|R'|}i^*}$, that is, the distance from $i_{|R'|}$ (the last customer in $R'$) to $i^*$. Finally, the length of $i_{R'}$ is given by $l_{i_{R'}}=\sum\limits_{r'=1}^{|R'|-1}{d_{i_{r'}i_{r'+1}}}$, that is, the sum of the distances between consecutive customers in $R'$.

\begin{figure}[!ht]
	\centering \includegraphics*[viewport = 150 190 470 730, height=\hsize]{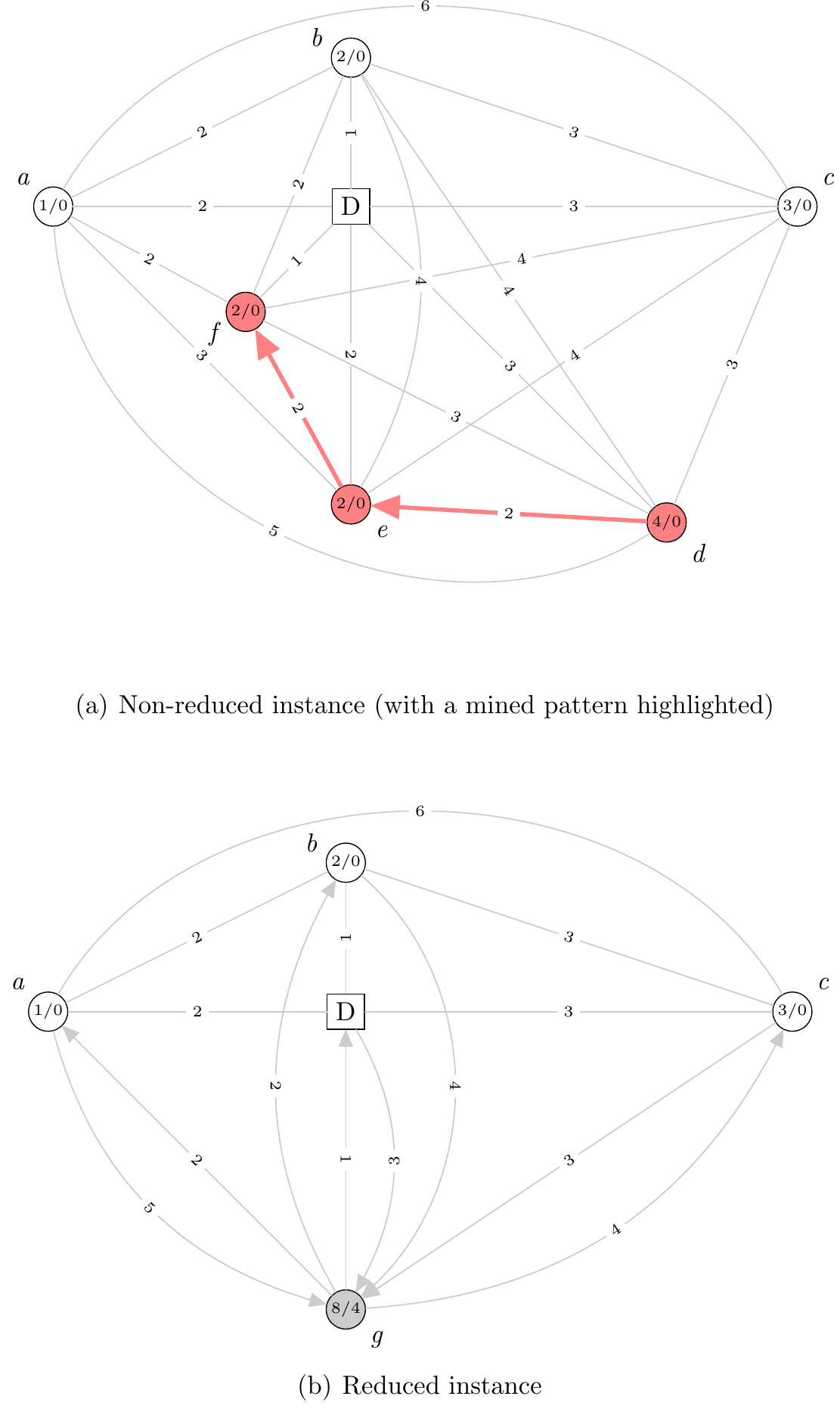}
	\caption{An example of reduction of an HFVRP instance based on a mined pattern}
	\label{fig:reduction}
\end{figure}

From the definition of a reduced instance described above, it is noted that, even if all distances $d_{ij}$ associated with the arcs $(i, j) \in A$ are symmetric for an instance of the HFVRP (i.e., $d_{ij}=d_{ji}, \forall i,j \in V$), a reduced version will present asymmetric distances associated with the arcs in $A^*$ that connect customer cluster vertices to the other vertices in $V^*$. This detail is relevant because many of the approaches proposed for the HFVRP handle only instances with symmetric distances or treat symmetric and asymmetric instances differently. Fig. \ref{fig:reduction} shows the reduction of an HFVRP instance based on a mined pattern.

Fig. \ref{fig:reduction}(a) presents a representation of the graph $G = (V, A)$ in the extended model described above. In this representation, the circular vertices (customers) present, in addition to the value of the demand, the value of the length (to the right) of the corresponding customer. In this non-reduced instance, all customer vertices have length $0$. The pattern used in the presented reduction is highlighted in the graph of Fig.~\ref{fig:reduction}(a). This pattern consists of a single route segment: $(R'_1, u_1)$, where $R'_1 = (d, e, f)$. The graph $G^* = (V^*, A^*)$ associated with the reduced version of the instance is presented in Fig. \ref{fig:reduction}(b). The customer vertices in $R'_1$ are replaced in the graph by a customer cluster represented by the vertex $g$. This customer cluster's demand $q_g$ is given by the sum of the demands of the customers it replaces: $q_g=q_d+q_e+q_f=4+2+2=8$. The distance from each of the other vertices to $g$ is given by the distances from these vertices to $d$ (first customer in $R'_1$): $d_{Dg}=d_{Dd}=3$, $d_{ag}=d_{ad}=5$, $d_{bg}=d_{bd}=4$ and $d_{cg}=d_{cd}=3$. The distance from $g$ to each of the other vertices is given by the distances from $f$ (last customer in $R'_1$) to these vertices: $d_{gD}=d_{fD}=1$, $d_{ga}=d_{fa}=2$, $d_{gb}=d_{fb}=2$ and $d_{gc}=d_{fc}=4$. Finally, the length $l_g$ of $g$ is given by the sum of the distances between consecutive customers in $R'_1$: $l_g=d_{de}+d_{ef}=2+2=4$.

\section{Computational results} \label{sec:results}

This section reports the results of the experiments carried out. The MineReduce-based heuristic, called MineReduce-MS-ILS (MineReduce for short), is evaluated and compared to the MS-ILS heuristic proposed by \citet{pennaetal2013a} and to its previous hybrid data mining version, MDM-MS-ILS \citep{maiaetal2018}.

In these experiments, we used the 96 HFFVRP-FD instances described by \citet{duhameletal2010}, which are divided into four sets: Set 1, with 15 instances, each with fewer than 100 customers; Set 2, with 38 instances, each with 100 to 150 customers; Set 3, with 31 instances, each with 151 to 200 customers; and Set 4, with 12 instances, each with more than 200 customers.

The MineReduce-MS-ILS heuristic was implemented based on the original source code for the MS-ILS heuristic. All of the three heuristics were compiled using the GCC C++ compiler (g++) version 4.8.2. The experiments were run on an Intel\textsuperscript{\textregistered} Core\textsuperscript{TM} i7-5500U 2.40 GHz CPU with 8 GB of RAM running Windows 10 (x64). Each heuristic was executed 20 times, using distinct random number seeds, for each instance. In each case, we present the following values obtained over the 20 runs: best solution cost, average solution cost and average computational time (in seconds).

Additionally, we have compared the results obtained by MineReduce to those obtained by the state-of-the-art algorithms of \citet{kochetov_khmelev2015} and \citet{pennaetal2019}. In this comparison, we have considered their reported results since we did not have the chance to run experiments with their algorithms.

We report the parameters setting in Section \ref{subsec:param}. Section \ref{subsec:resultsHFFVRP} presents the main results, with comprehensive comparisons of the performance of the heuristics tested, as well as comparisons between the results obtained by MineReduce and the results reported in the literature for other state-of-the-art algorithms. Lastly, we present the results of additional experiments carried out to further inspect the behavior of the tested heuristics in Section~\ref{sec:behavior}.

\subsection{Parameters setting} \label{subsec:param}

The MS-ILS heuristic has only two parameters, both related to stopping criteria: $MaxIter$, which is the number of multi-start iterations, and $\beta$, which is used in the calculation of the maximum number of consecutive perturbations allowed without improvements in the ILS ($MaxIterILS=n+\beta v$). It has been previously observed that the quality of the solutions and the computational time tend to increase as the values of $MaxIter$ and $\beta$ are increased~\citep{pennaetal2013a}, which was expected since they define the number of trials given to the algorithm. Therefore, the question of choosing values for these parameters is a trade-off between solution quality and computational time. In these experiments, we have adopted the following values for these parameters (for all of the heuristics): $MaxIter = 100$, which was adopted by~\citet{maiaetal2018}, and $\beta = 5$, which was recommended by~\citet{pennaetal2013a}.

For tuning the other parameters of the hybrid heuristics (MDM-MS-ILS and MineReduce), related to the data mining procedure, we have used the irace package, which is a software package that implements iterated racing procedures that have been successfully used to configure various state-of-the-art algorithms~\citep{lopez-ibanezetal2016}. Specifically, we have used irace to derive a set of appropriate parameter values for each heuristic. The list of training instances (which have been taken out of the comparisons in Section~\ref{subsec:resultsHFFVRP}) was composed of the first instance of each set -- i.e., instances 01, 03, 02 and 18 from Sets 1, 2, 3 and 4 of~\citet{duhameletal2010}, respectively. We have defined as the stopping criterion for irace a maximum total execution time (irace's maxTime parameter) of 45 hours. Besides the stopping criterion, we have used irace's default configuration.

Table~\ref{tab:paramtuning} presents the results of the tuning process. The first three columns identify the parameters, their descriptions, and the sets of candidate values provided to irace, respectively. The remaining columns present the best configurations found by irace for each heuristic, which we have used in the experiments reported in this work.

\begin{table}[!ht] \footnotesize
\begin{center}
\caption{Parameters tuning for MDM-MS-ILS and MineReduce with the irace package}
\label{tab:paramtuning}
\begin{tabular}{lllcc}
\hline
Parameter & Description & Tested values & MDM-MS-ILS & MineReduce \\ \hline
$d$ & Maximum size of the elite set & $\{10, 15, 20\}$ & 10 & 10 \\
$MaxP$ & Maximum size of the patterns set & $\{6..10\}$ & 9 & 6 \\
$MinSup$ & Minimum support & $\{0.2, 0.3, ..., 1.0\}$ & 0.7 & 0.2 \\
$\delta$ & \begin{tabular}[c]{@{}l@{}}Number of iterations without modification\\in the elite set for it to be considered stable\end{tabular} & $\{3..7\}$ & 3 & 3 \\ \hline
\end{tabular}
\end{center}
\end{table}

\subsection{Main results} \label{subsec:resultsHFFVRP}

This section presents the main results obtained in our computational experiments. For each set of instances, one table presents the results as follows. There is one row for each instance (I), containing the best known solution (BKS) cost reported in the literature -- with the indication of the references reporting it (letter marks) -- and the results achieved by the compared heuristics. Two extra rows are included at the bottom. The first indicates the number of times each heuristic attained the best result (\# of wins), whereas the second indicates the average percentage difference (APD) obtained by the hybrid heuristics -- i.e., the mean of the percentage differences in average cost or time for each of the hybrid heuristics compared to the MS-ILS heuristic. In the comparisons, the best values among all heuristics are in boldface. Costs equal to the BKS are shown in italic and the new best solutions found (i.e., solutions that are better than the previously best known solutions) are underlined.

Moreover, we present an assessment of the statistical significance of the variations in the average costs achieved by the compared heuristics. For this evaluation, we have used a one-tailed paired Student's t-test per instance for each pair of heuristics (MS-ILS vs. MineReduce and MDM-MS-ILS vs. MineReduce), with a significance level of $5\%$. The statistically significant differences are indicated, for each instance, in the average cost column of the heuristic that has the advantage.

Table \ref{tab:HFFVRPDuh1} presents the comparison of the results for the instances from Set 1 of \citet{duhameletal2010}. For this set, composed of small instances, the overall results were balanced regarding solution quality, with small differences between the heuristics. Regarding the average times, on the other hand, MineReduce obtained the best values for all instances and the best, and expressive, APD (-63.72\%). A new best solution to instance 39 was found by all heuristics.

\begin{table}[!ht] \scriptsize
\setlength{\tabcolsep}{5pt}
\begin{center}
\caption{Results -- Set 1 of \citet{duhameletal2010}}
\label{tab:HFFVRPDuh1}
\begin{threeparttable}
\begin{tabular}{rrrrrrrrrrrrr}
\hline
\multicolumn{1}{c}{} & \multicolumn{1}{c}{} & \multicolumn{3}{c}{MS-ILS} & \multicolumn{1}{c}{} & \multicolumn{3}{c}{MDM-MS-ILS} & \multicolumn{1}{c}{} & \multicolumn{3}{c}{MineReduce} \\ \cline{3-5} \cline{7-9} \cline{11-13} 
\multicolumn{1}{c}{I} & \multicolumn{1}{c}{BKS} & \multicolumn{1}{c}{\begin{tabular}[c]{@{}c@{}}Best\\ Cost\end{tabular}} & \multicolumn{1}{c}{\begin{tabular}[c]{@{}c@{}}Avg.\\ Cost\end{tabular}} & \multicolumn{1}{c}{\begin{tabular}[c]{@{}c@{}}Avg.\\ Time\end{tabular}} & \multicolumn{1}{c}{} & \multicolumn{1}{c}{\begin{tabular}[c]{@{}c@{}}Best\\ Cost\end{tabular}} & \multicolumn{1}{c}{\begin{tabular}[c]{@{}c@{}}Avg.\\ Cost\end{tabular}} & \multicolumn{1}{c}{\begin{tabular}[c]{@{}c@{}}Avg.\\ Time\end{tabular}} & \multicolumn{1}{c}{} & \multicolumn{1}{c}{\begin{tabular}[c]{@{}c@{}}Best\\ Cost\end{tabular}} & \multicolumn{1}{c}{\begin{tabular}[c]{@{}c@{}}Avg.\\ Cost\end{tabular}} & \multicolumn{1}{c}{\begin{tabular}[c]{@{}c@{}}Avg.\\ Time\end{tabular}} \\ \hline
08 & 4591.75\tnote{c--f} & 4594.07 & 4596.13 & 52.1 &  & \textit{\textbf{4591.75}} & 4595.89 & 44.6 &  & \textit{\textbf{4591.75}} & \textbf{4594.20}\tnote{\dag\ddag} & \textbf{30.2} \\
10 & 2107.55\tnote{a--f} & \textit{\textbf{2107.55}} & 2107.58 & 68.1 &  & \textit{\textbf{2107.55}} & \textit{\textbf{2107.55}} & 56.7 &  & \textit{\textbf{2107.55}} & \textit{\textbf{2107.55}} & \textbf{29.5} \\
11 & 3367.41\tnote{d--g} & 3368.50 & 3375.86 & 101.7 &  & 3368.50 & \textbf{3374.01} & 84.6 &  & \textit{\textbf{3367.41}} & 3376.12 & \textbf{49.0} \\
36 & 5684.61\tnote{de} & \textbf{5684.62} & 5704.48 & 177.8 &  & \textbf{5684.62} & 5703.50 & 153.7 &  & \textbf{5684.62} & \textbf{5702.82} & \textbf{70.5} \\
39 & 2921.36\tnote{f} & {\ul \textbf{2920.93}} & \textbf{2931.17}\tnote{\dag} & 116.5 &  & {\ul \textbf{2920.93}} & 2933.49 & 104.6 &  & {\ul \textbf{2920.93}} & 2933.75 & \textbf{47.1} \\
43 & 8737.02\tnote{c--e} & 8744.50 & 8754.30 & 119.9 &  & \textit{\textbf{8737.02}} & \textbf{8751.34}\tnote{\ddag} & 105.7 &  & 8739.36 & 8756.97 & \textbf{55.7} \\
52 & 4027.27\tnote{d--f} & \textbf{4029.42} & \textbf{4029.42}\tnote{\dag} & 35.7 &  & \textbf{4029.42} & \textbf{4029.42}\tnote{\ddag} & 31.0 &  & \textbf{4029.42} & 4030.35 & \textbf{15.6} \\
55 & 10244.34\tnote{a--f} & \textit{\textbf{10244.34}} & \textbf{10255.81}\tnote{\dag} & 24.5 &  & \textit{\textbf{10244.34}} & 10258.72\tnote{\ddag} & 21.2 &  & \textit{\textbf{10244.34}} & 10313.29 & \textbf{14.0} \\
70 & 6684.56\tnote{ef} & 6688.69 & 6698.32 & 57.3 &  & \textbf{6685.24} & \textbf{6693.99} & 49.9 &  & \textbf{6685.24} & 6695.22 & \textbf{29.8} \\
75 & 452.85\tnote{a--g} & \textit{\textbf{452.85}} & \textit{\textbf{452.85}} & 4.2 &  & \textit{\textbf{452.85}} & \textit{\textbf{452.85}} & 3.4 &  & \textit{\textbf{452.85}} & \textit{\textbf{452.85}} & \textbf{3.2} \\
82 & 4766.74\tnote{c--f} & \textit{\textbf{4766.74}} & 4771.10 & 49.6 &  & \textit{\textbf{4766.74}} & 4771.26 & 46.7 &  & \textit{\textbf{4766.74}} & \textbf{4770.34} & \textbf{30.0} \\
92 & 564.39\tnote{a--g} & \textit{\textbf{564.39}} & \textit{\textbf{564.39}} & 14.4 &  & \textit{\textbf{564.39}} & \textit{\textbf{564.39}} & 12.4 &  & \textit{\textbf{564.39}} & \textit{\textbf{564.39}} & \textbf{9.1} \\
93 & 1036.99\tnote{b--g} & \textit{\textbf{1036.99}} & \textbf{1037.18}\tnote{\dag} & 15.5 &  & \textit{\textbf{1036.99}} & 1037.28\tnote{\ddag} & 13.4 &  & \textit{\textbf{1036.99}} & 1038.38 & \textbf{8.6} \\
94 & 1378.25\tnote{a--g} & \textit{\textbf{1378.25}} & \textit{\textbf{1378.25}} & 35.6 &  & \textit{\textbf{1378.25}} & 1378.27 & 31.6 &  & \textit{\textbf{1378.25}} & 1378.34 & \textbf{17.1} \\ \hline
\multicolumn{2}{l}{\# of wins} & 10 & \textbf{7} & - &  & \textbf{13} & \textbf{7} & - &  & \textbf{13} & 6 & \textbf{14} \\
\multicolumn{2}{l}{APD} &  &  &  &  & \textbf{-0.01\%} & \textbf{0.00\%} & -14.28\% &  & \textbf{-0.01\%} & 0.05\% & \textbf{-63.72\%} \\ \hline
\end{tabular}
\begin{tablenotes}
\item [\dag]Statistically significant (MS-ILS vs. MineReduce)
\item [\ddag]Statistically significant (MDM-MS-ILS vs. MineReduce)
\item [a]\citet{duhameletal2010}
\item [b]\citet{duhameletal2011}
\item [c]\citet{maiaetal2018}
\item [d]\citet{duhameletal2013}
\item [e]\citet{pennaetal2019}
\item [f]\citet{pennaetal2013b}
\item [g]\citet{kochetov_khmelev2015}
\end{tablenotes}
\end{threeparttable}
\end{center}
\end{table}

Table \ref{tab:HFFVRPDuh2} presents the comparison of the results for the instances from Set 2 of \citet{duhameletal2010}. MineReduce outperformed the other heuristics, achieving the best average costs for 26 of the 37 instances, with an APD of -0.19\%. Its advantage was statistically significant for 22 instances in comparison to MS-ILS and for 19 instances in comparison to MDM-MS-ILS. MineReduce found new best solutions to five instances and the best known solutions to 12 other instances. Furthermore, MineReduce had the best average times for all instances, with the expressive APD of -65.09\%.

\begin{table}[!ht] \scriptsize
\setlength{\tabcolsep}{5pt}
\begin{center}
\caption{Results -- Set 2 of \citet{duhameletal2010}}
\label{tab:HFFVRPDuh2}
\begin{threeparttable}
\begin{tabular}{rrrrrrrrrrrrr}
\hline
\multicolumn{1}{c}{} & \multicolumn{1}{c}{} & \multicolumn{3}{c}{MS-ILS} & \multicolumn{1}{c}{} & \multicolumn{3}{c}{MDM-MS-ILS} & \multicolumn{1}{c}{} & \multicolumn{3}{c}{MineReduce} \\ \cline{3-5} \cline{7-9} \cline{11-13} 
\multicolumn{1}{c}{I} & \multicolumn{1}{c}{BKS} & \multicolumn{1}{c}{\begin{tabular}[c]{@{}c@{}}Best\\ Cost\end{tabular}} & \multicolumn{1}{c}{\begin{tabular}[c]{@{}c@{}}Avg.\\ Cost\end{tabular}} & \multicolumn{1}{c}{\begin{tabular}[c]{@{}c@{}}Avg.\\ Time\end{tabular}} & \multicolumn{1}{c}{} & \multicolumn{1}{c}{\begin{tabular}[c]{@{}c@{}}Best\\ Cost\end{tabular}} & \multicolumn{1}{c}{\begin{tabular}[c]{@{}c@{}}Avg.\\ Cost\end{tabular}} & \multicolumn{1}{c}{\begin{tabular}[c]{@{}c@{}}Avg.\\ Time\end{tabular}} & \multicolumn{1}{c}{} & \multicolumn{1}{c}{\begin{tabular}[c]{@{}c@{}}Best\\ Cost\end{tabular}} & \multicolumn{1}{c}{\begin{tabular}[c]{@{}c@{}}Avg.\\ Cost\end{tabular}} & \multicolumn{1}{c}{\begin{tabular}[c]{@{}c@{}}Avg.\\ Time\end{tabular}} \\ \hline
05 & 10876.48\tnote{d} & 10904.08 & 10973.10 & 129.7 &  & 10909.01 & 10950.87 & 110.8 &  & \textit{\textbf{10876.48}} & \textbf{10925.52}\tnote{\dag\ddag} & \textbf{76.4} \\
06 & 11688.64\tnote{d} & 11771.85 & 11823.31 & 186.3 &  & 11723.12 & 11806.50 & 160.8 &  & \textbf{11696.83} & \textbf{11761.32}\tnote{\dag\ddag} & \textbf{125.0} \\
07 & 8074.64\tnote{f} & 8089.72 & 8149.27 & 122.0 &  & 8088.02 & 8141.89 & 105.0 &  & \textbf{8076.53} & \textbf{8129.35} & \textbf{68.6} \\
12 & 3543.99\tnote{a--f} & \textit{\textbf{3543.99}} & 3546.89 & 175.9 &  & \textit{\textbf{3543.99}} & \textbf{3546.18} & 151.0 &  & \textit{\textbf{3543.99}} & 3547.65 & \textbf{82.9} \\
13 & 6696.43\tnote{b--d} & 6708.49 & 6714.25 & 256.3 &  & 6701.58 & 6714.60 & 219.7 &  & \textbf{6697.58} & \textbf{6706.94}\tnote{\dag\ddag} & \textbf{102.8} \\
16 & 4156.97\tnote{b--f} & \textit{\textbf{4156.97}} & 4163.69 & 253.4 &  & \textit{\textbf{4156.97}} & \textbf{4162.15} & 220.7 &  & \textit{\textbf{4156.97}} & 4163.04 & \textbf{126.1} \\
17 & 5362.83\tnote{bd} & \textbf{5365.52} & 5380.09 & 120.9 &  & 5367.49 & \textbf{5376.80} & 103.0 &  & 5365.94 & 5381.80 & \textbf{61.3} \\
2A & 7793.16\tnote{b--e} & \textit{\textbf{7793.16}} & 7816.48 & 241.1 &  & \textit{\textbf{7793.16}} & \textbf{7807.30} & 210.2 &  & \textit{\textbf{7793.16}} & 7809.81 & \textbf{113.1} \\
2B & 8462.56\tnote{d} & 8473.84 & 8497.01 & 304.6 &  & 8476.92 & 8502.85 & 260.1 &  & {\ul \textbf{8453.35}} & \textbf{8477.33}\tnote{\dag\ddag} & \textbf{137.0} \\
21 & 5139.84\tnote{cd} & 5148.03 & \textbf{5160.73} & 245.1 &  & \textit{\textbf{5139.84}} & 5162.26 & 214.0 &  & \textit{\textbf{5139.84}} & 5162.74 & \textbf{110.3} \\
25 & 7206.64\tnote{b} & \textbf{7209.29} & 7250.69 & 537.7 &  & 7236.75 & 7252.06 & 482.2 &  & 7209.50 & \textbf{7231.23}\tnote{\dag\ddag} & \textbf{281.3} \\
26 & 6393.47\tnote{f} & 6448.79 & 6458.54 & 752.0 &  & \textbf{6418.40} & \textbf{6456.43} & 668.2 &  & 6438.69 & 6458.32 & \textbf{275.7} \\
28 & 5530.55\tnote{d} & 5531.30 & 5540.40 & 378.7 &  & 5535.15 & 5541.76 & 322.5 &  & {\ul \textbf{5529.05}} & \textbf{5537.72}\tnote{\dag\ddag} & \textbf{171.4} \\
30 & 6313.39\tnote{b} & \textbf{6320.14} & \textbf{6343.28}\tnote{\dag} & 338.5 &  & 6333.82 & 6346.37 & 300.5 &  & 6331.40 & 6350.05 & \textbf{117.9} \\
31 & 4091.52\tnote{bd} & 4107.54 & 4125.63 & 519.0 &  & 4097.97 & 4122.48 & 441.9 &  & \textit{\textbf{4091.52}} & \textbf{4113.60}\tnote{\dag} & \textbf{218.8} \\
34 & 5747.25\tnote{d} & 5778.46 & 5815.53 & 350.6 &  & 5775.58 & 5815.46 & 300.6 &  & \textbf{5765.08} & \textbf{5792.18}\tnote{\dag\ddag} & \textbf{185.6} \\
40 & 11118.57\tnote{d} & 11129.37 & 11163.23 & 300.9 &  & 11136.31 & 11160.28 & 264.6 &  & {\ul \textbf{11111.89}} & \textbf{11140.70}\tnote{\dag\ddag} & \textbf{162.0} \\
41 & 7571.44\tnote{e} & 7619.76 & 7715.74 & 312.6 &  & 7619.76 & 7714.90 & 279.9 &  & \textbf{7573.24} & \textbf{7637.37}\tnote{\dag\ddag} & \textbf{204.9} \\
47 & 16156.12\tnote{d} & 16243.28 & 16314.30 & 151.4 &  & 16232.57 & 16299.40 & 134.4 &  & \textit{\textbf{16156.12}} & \textbf{16263.38}\tnote{\dag\ddag} & \textbf{83.9} \\
48 & 21287.90\tnote{e} & \textit{\textbf{21287.90}} & 21451.91 & 194.7 &  & 21383.96 & 21463.82 & 167.4 &  & 21329.71 & \textbf{21413.41}\tnote{\dag\ddag} & \textbf{99.0} \\
51 & 7721.47\tnote{bd} & \textbf{7769.42} & \textbf{7794.11} & 283.8 &  & 7780.04 & 7794.51\tnote{\ddag} & 245.8 &  & 7780.04 & 7804.28 & \textbf{117.0} \\
53 & 6434.83\tnote{b--e} & \textit{\textbf{6434.83}} & 6455.63 & 158.8 &  & \textit{\textbf{6434.83}} & 6458.59 & 138.6 &  & \textit{\textbf{6434.83}} & \textbf{6448.50}\tnote{\dag\ddag} & \textbf{80.4} \\
60 & 17036.59\tnote{d} & 17054.68 & 17101.45 & 247.5 &  & 17051.56 & 17090.47 & 212.9 &  & \textbf{17045.33} & \textbf{17082.39}\tnote{\dag} & \textbf{125.2} \\
61 & 7292.03\tnote{cd} & 7294.03 & 7304.63 & 200.6 &  & \textit{\textbf{7292.03}} & \textbf{7302.24}\tnote{\ddag} & 181.5 &  & \textit{\textbf{7292.03}} & 7305.89 & \textbf{96.7} \\
66 & 12783.94\tnote{d} & 12809.36 & 12882.42 & 646.0 &  & 12834.58 & 12889.40 & 465.2 &  & {\ul \textbf{12772.07}} & \textbf{12839.71}\tnote{\dag\ddag} & \textbf{295.9} \\
68 & 8935.89\tnote{c} & 8997.40 & 9079.65 & 268.8 &  & 8991.80 & 9066.08 & 203.0 &  & {\ul \textbf{8919.16}} & \textbf{8992.62}\tnote{\dag\ddag} & \textbf{153.9} \\
73 & 10195.13\tnote{f} & 10204.77 & 10212.39 & 281.2 &  & 10204.77 & 10213.31 & 235.0 &  & \textbf{10203.84} & \textbf{10209.61}\tnote{\dag\ddag} & \textbf{136.7} \\
74 & 11586.58\tnote{d} & 11595.36 & 11608.81 & 299.2 &  & \textbf{11586.87} & 11609.91 & 258.0 &  & \textbf{11586.87} & \textbf{11599.16}\tnote{\dag\ddag} & \textbf{150.6} \\
79 & 7259.54\tnote{bd} & 7275.74 & 7307.53 & 580.0 &  & 7262.91 & 7292.31 & 493.6 &  & \textbf{7262.02} & \textbf{7290.96}\tnote{\dag} & \textbf{234.8} \\
81 & 10675.92\tnote{f} & \textbf{10689.90} & 10702.57 & 181.7 &  & 10693.70 & 10706.12 & 157.3 &  & 10693.70 & \textbf{10699.79}\tnote{\ddag} & \textbf{95.1} \\
83 & 10019.15\tnote{b} & 10041.08 & 10052.44 & 219.2 &  & 10029.79 & 10047.39 & 198.3 &  & \textit{\textbf{10019.15}} & \textbf{10046.97} & \textbf{124.7} \\
84 & 7227.88\tnote{bd} & 7237.13 & 7262.42 & 125.3 &  & \textit{\textbf{7227.88}} & \textbf{7258.76}\tnote{\ddag} & 118.6 &  & \textit{\textbf{7227.88}} & 7267.13 & \textbf{72.4} \\
85 & 8773.08\tnote{d} & \textbf{8825.54} & 8863.32 & 257.5 &  & 8827.98 & 8862.83 & 250.9 &  & 8827.98 & \textbf{8857.80} & \textbf{171.6} \\
87 & 3753.87\tnote{a--f} & \textit{\textbf{3753.87}} & 3757.77 & 127.9 &  & \textit{\textbf{3753.87}} & 3757.13 & 121.9 &  & \textit{\textbf{3753.87}} & \textbf{3755.06}\tnote{\dag} & \textbf{66.2} \\
88 & 12388.23\tnote{d} & 12429.11 & 12512.64 & 156.9 &  & 12429.11 & 12494.25 & 153.4 &  & \textbf{12402.85} & \textbf{12447.06}\tnote{\dag\ddag} & \textbf{118.6} \\
89 & 7086.36\tnote{d} & 7105.15 & 7128.66 & 245.1 &  & 7100.90 & 7126.66 & 233.5 &  & \textbf{7095.33} & \textbf{7110.97}\tnote{\dag\ddag} & \textbf{140.0} \\
90 & 2346.13\tnote{b} & 2347.81 & 2358.88 & 105.3 &  & \textbf{2346.43} & \textbf{2357.25} & 87.0 &  & \textbf{2346.43} & 2357.59 & \textbf{54.8} \\ \hline
\multicolumn{2}{l}{\# of wins} & 12 & 3 & - &  & 11 & 8 & - &  & \textbf{29} & \textbf{26} & \textbf{37} \\
\multicolumn{2}{l}{APD} &  &  &  &  & -0.01\% & -0.04\% & -13.71\% &  & \textbf{-0.15\%} & \textbf{-0.19\%} & \textbf{-65.09\%} \\ \hline
\end{tabular}
\begin{tablenotes}
\item [\dag]Statistically significant (MS-ILS vs. MineReduce)
\item [\ddag]Statistically significant (MDM-MS-ILS vs. MineReduce)
\item [a]\citet{duhameletal2010}
\item [b]\citet{duhameletal2013}
\item [c]\citet{pennaetal2013b}
\item [d]\citet{pennaetal2019}
\item [e]\citet{maiaetal2018}
\item [f]\citet{kochetov_khmelev2015}
\end{tablenotes}
\end{threeparttable}
\end{center}
\end{table}

Table \ref{tab:HFFVRPDuh3} presents the comparison of the results for the instances from Set 3 of \citet{duhameletal2010}. MineReduce again outperformed the other heuristics, achieving the best average costs for all 30 instances, with an APD of -0.46\%. The differences were statistically significant for 27 instances in comparison to MS-ILS, and for 29 instances in comparison to MDM-MS-ILS. MineReduce found new best solutions to nine instances and the best known solution to another one. MineReduce also obtained the best average times for all instances, with an APD of -57.01\%.

\begin{table}[!ht] \scriptsize
\setlength{\tabcolsep}{5pt}
\begin{center}
\caption{Results -- Set 3 of \citet{duhameletal2010}}
\label{tab:HFFVRPDuh3}
\begin{threeparttable}
\begin{tabular}{rrrrrrrrrrrrr}
\hline
\multicolumn{1}{c}{} & \multicolumn{1}{c}{} & \multicolumn{3}{c}{MS-ILS} & \multicolumn{1}{c}{} & \multicolumn{3}{c}{MDM-MS-ILS} & \multicolumn{1}{c}{} & \multicolumn{3}{c}{MineReduce} \\ \cline{3-5} \cline{7-9} \cline{11-13} 
\multicolumn{1}{c}{I} & \multicolumn{1}{c}{BKS} & \multicolumn{1}{c}{\begin{tabular}[c]{@{}c@{}}Best\\ Cost\end{tabular}} & \multicolumn{1}{c}{\begin{tabular}[c]{@{}c@{}}Avg.\\ Cost\end{tabular}} & \multicolumn{1}{c}{\begin{tabular}[c]{@{}c@{}}Avg.\\ Time\end{tabular}} & \multicolumn{1}{c}{} & \multicolumn{1}{c}{\begin{tabular}[c]{@{}c@{}}Best\\ Cost\end{tabular}} & \multicolumn{1}{c}{\begin{tabular}[c]{@{}c@{}}Avg.\\ Cost\end{tabular}} & \multicolumn{1}{c}{\begin{tabular}[c]{@{}c@{}}Avg.\\ Time\end{tabular}} & \multicolumn{1}{c}{} & \multicolumn{1}{c}{\begin{tabular}[c]{@{}c@{}}Best\\ Cost\end{tabular}} & \multicolumn{1}{c}{\begin{tabular}[c]{@{}c@{}}Avg.\\ Cost\end{tabular}} & \multicolumn{1}{c}{\begin{tabular}[c]{@{}c@{}}Avg.\\ Time\end{tabular}} \\ \hline
04 & 10748.17\tnote{e} & 10792.86 & 10827.34 & 550.9 &  & 10772.57 & 10820.86 & 470.5 &  & {\ul \textbf{10725.96}} & \textbf{10772.83}\tnote{\dag\ddag} & \textbf{329.2} \\
09 & 7603.38\tnote{e} & 7651.75 & 7667.25 & 369.0 &  & 7648.92 & 7663.31 & 313.8 &  & \textbf{7618.34} & \textbf{7652.09}\tnote{\dag\ddag} & \textbf{225.6} \\
14 & 5644.98\tnote{a} & 5663.21 & 5705.01 & 672.8 &  & 5666.74 & 5703.68 & 602.6 &  & \textbf{5654.49} & \textbf{5671.42}\tnote{\dag\ddag} & \textbf{449.5} \\
15 & 8220.64\tnote{e} & 8283.50 & 8313.50 & 505.0 &  & 8281.52 & 8315.08 & 443.2 &  & \textbf{8229.06} & \textbf{8266.88}\tnote{\dag\ddag} & \textbf{412.0} \\
24 & 9101.47\tnote{a} & 9170.10 & 9221.76 & 639.4 &  & 9163.03 & 9219.39 & 519.8 &  & \textbf{9128.64} & \textbf{9169.28}\tnote{\dag\ddag} & \textbf{373.5} \\
29 & 9142.86\tnote{be} & {\ul 9140.34} & 9154.66 & 850.1 &  & {\ul 9139.50} & 9158.06 & 728.7 &  & {\ul \textbf{9136.41}} & \textbf{9151.44}\tnote{\ddag} & \textbf{275.4} \\
33 & 9410.99\tnote{e} & 9461.90 & 9494.98 & 824.2 &  & 9472.89 & 9501.33 & 731.5 &  & \textbf{9411.12} & \textbf{9440.74}\tnote{\dag\ddag} & \textbf{505.4} \\
35 & 9555.92\tnote{e} & 9580.51 & 9640.73 & 394.3 &  & 9594.32 & 9637.08 & 348.2 &  & \textbf{9565.36} & \textbf{9617.15}\tnote{\dag\ddag} & \textbf{241.2} \\
37 & 6850.77\tnote{e} & 6869.39 & 6887.86 & 605.9 &  & 6866.77 & 6890.96 & 525.1 &  & \textbf{6854.99} & \textbf{6869.01}\tnote{\dag\ddag} & \textbf{345.9} \\
42 & 10817.90\tnote{e} & 10931.85 & 11057.02 & 1069.6 &  & 10960.76 & 11045.49 & 956.7 &  & \textbf{10842.11} & \textbf{10918.29}\tnote{\dag\ddag} & \textbf{767.3} \\
44 & 12191.48\tnote{e} & 12241.46 & 12353.84 & 743.4 &  & 12243.66 & 12371.33 & 632.4 &  & \textbf{12214.60} & \textbf{12308.14}\tnote{\ddag} & \textbf{331.0} \\
45 & 10476.25\tnote{e} & 10512.17 & 10627.05 & 581.9 &  & 10512.17 & 10624.44 & 501.6 &  & \textbf{10497.86} & \textbf{10540.84}\tnote{\dag\ddag} & \textbf{343.1} \\
50 & 12370.94\tnote{e} & 12414.86 & 12455.66 & 1911.1 &  & 12405.04 & 12454.76 & 1703.6 &  & {\ul \textbf{12333.66}} & \textbf{12394.94}\tnote{\dag\ddag} & \textbf{836.6} \\
54 & 10351.97\tnote{e} & 10354.94 & 10458.09 & 772.8 &  & 10354.94 & 10467.68 & 687.2 &  & {\ul \textbf{10342.62}} & \textbf{10408.42}\tnote{\dag\ddag} & \textbf{377.3} \\
56 & 31030.19\tnote{e} & 31104.99 & 31262.19 & 400.9 &  & 31138.93 & 31267.46 & 358.4 &  & {\ul \textbf{31020.27}} & \textbf{31122.92}\tnote{\dag\ddag} & \textbf{280.9} \\
57 & 43378.37\tnote{c} & 44856.76 & 45012.38 & 475.2 &  & 44856.76 & 45051.96 & 413.3 &  & \textbf{44771.71} & \textbf{44854.99}\tnote{\dag\ddag} & \textbf{288.1} \\
59 & 14299.28\tnote{c} & 14322.92 & 14388.71 & 1156.0 &  & 14325.27 & 14386.27 & 1019.9 &  & \textbf{14310.58} & \textbf{14352.36}\tnote{\dag\ddag} & \textbf{511.9} \\
63 & 19951.76\tnote{a} & 20149.87 & 20297.18 & 461.6 &  & 20161.32 & 20299.49 & 450.9 &  & {\ul \textbf{19946.64}} & \textbf{20047.97}\tnote{\dag\ddag} & \textbf{347.2} \\
64 & 17135.16\tnote{be} & \textit{\textbf{17135.16}} & 17169.00 & 501.2 &  & 17154.96 & 17170.87 & 454.9 &  & \textit{\textbf{17135.16}} & \textbf{17162.28}\tnote{\ddag} & \textbf{222.0} \\
67 & 10850.16\tnote{d} & 10928.97 & 10982.48 & 898.8 &  & 10906.31 & 10982.45 & 601.3 &  & \textbf{10861.53} & \textbf{10935.95}\tnote{\dag\ddag} & \textbf{361.8} \\
69 & 9147.54\tnote{e} & 9177.52 & 9234.53 & 329.2 &  & 9177.52 & 9230.02 & 282.2 &  & {\ul \textbf{9143.84}} & \textbf{9190.45}\tnote{\dag\ddag} & \textbf{204.8} \\
71 & 9834.40\tnote{e} & 9928.42 & 9962.99 & 484.6 &  & 9890.50 & 9960.87 & 418.8 &  & \textbf{9847.58} & \textbf{9894.85}\tnote{\dag\ddag} & \textbf{253.6} \\
72 & 5883.33\tnote{b} & 5929.17 & 5956.65 & 856.0 &  & 5929.17 & 5955.71 & 751.0 &  & \textbf{5895.89} & \textbf{5929.19}\tnote{\dag\ddag} & \textbf{469.2} \\
76 & 11994.40\tnote{e} & 12031.65 & 12088.63 & 464.6 &  & 12055.86 & 12089.98 & 395.1 &  & \textbf{12009.05} & \textbf{12040.47}\tnote{\dag\ddag} & \textbf{283.4} \\
77 & 6916.01\tnote{e} & 6947.64 & 6992.01 & 1529.3 &  & 6932.97 & 6982.48 & 1304.8 &  & \textbf{6920.92} & \textbf{6959.31}\tnote{\dag\ddag} & \textbf{639.0} \\
78 & 7035.01\tnote{a} & 7134.80 & 7146.32 & 1205.1 &  & \textbf{7056.50} & 7133.94 & 1122.3 &  & 7062.22 & \textbf{7123.49}\tnote{\dag} & \textbf{522.9} \\
80 & 6816.89\tnote{a} & 6834.00 & 6853.94 & 611.0 &  & 6829.59 & 6848.51 & 552.6 &  & \textbf{6826.59} & \textbf{6833.72}\tnote{\dag\ddag} & \textbf{296.6} \\
86 & 9027.84\tnote{e} & 9043.61 & 9066.53 & 401.9 &  & 9033.46 & 9064.96 & 390.3 &  & {\ul \textbf{9024.02}} & \textbf{9050.11}\tnote{\dag\ddag} & \textbf{236.1} \\
91 & 6374.01\tnote{c} & 6405.51 & 6423.51 & 629.3 &  & 6388.10 & 6421.60 & 527.3 &  & {\ul \textbf{6363.21}} & \textbf{6384.12}\tnote{\dag\ddag} & \textbf{487.1} \\
95 & 6175.62\tnote{be} & 6233.26 & 6239.71 & 458.3 &  & 6229.31 & 6239.54 & 387.0 &  & \textbf{6227.17} & \textbf{6233.96}\tnote{\dag\ddag} & \textbf{241.9} \\ \hline
\multicolumn{2}{l}{\# of wins} & 1 & - & - &  & 1 & - & - &  & \textbf{29} & \textbf{30} & \textbf{30} \\
\multicolumn{2}{l}{APD} &  &  &  &  & -0.06\% & -0.01\% & -13.83\% &  & \textbf{-0.39\%} & \textbf{-0.46\%} & \textbf{-57.01\%} \\ \hline
\end{tabular}
\begin{tablenotes}
\item [\dag]Statistically significant (MS-ILS vs. MineReduce)
\item [\ddag]Statistically significant (MDM-MS-ILS vs. MineReduce)
\item [a]\citet{duhameletal2013}
\item [b]\citet{pennaetal2013b}
\item [c]\citet{kochetov_khmelev2015}
\item [d]\citet{maiaetal2018}
\item [e]\citet{pennaetal2019}
\end{tablenotes}
\end{threeparttable}
\end{center}
\end{table}

Table \ref{tab:HFFVRPDuh4} presents the comparison of the results for the instances from Set 4 of \citet{duhameletal2010}. MineReduce once more outperformed the other heuristics, achieving the best average costs for all 11 instances, with an APD of -0.52\%. The differences were statistically significant in comparison to both MS-ILS and MDM-MS-ILS for all instances but one. Furthermore, MineReduce found new best solutions to six instances and obtained the best average times for all instances, with an APD of -53.24\%.

\begin{table}[!ht] \scriptsize
\setlength{\tabcolsep}{5pt}
\begin{center}
\caption{Results -- Set 4 of \citet{duhameletal2010}}
\label{tab:HFFVRPDuh4}
\begin{threeparttable}
\begin{tabular}{rrrrrrrrrrrrr}
\hline
\multicolumn{1}{c}{} & \multicolumn{1}{c}{} & \multicolumn{3}{c}{MS-ILS} & \multicolumn{1}{c}{} & \multicolumn{3}{c}{MDM-MS-ILS} & \multicolumn{1}{c}{} & \multicolumn{3}{c}{MineReduce} \\ \cline{3-5} \cline{7-9} \cline{11-13} 
\multicolumn{1}{c}{I} & \multicolumn{1}{c}{BKS} & \multicolumn{1}{c}{\begin{tabular}[c]{@{}c@{}}Best\\ Cost\end{tabular}} & \multicolumn{1}{c}{\begin{tabular}[c]{@{}c@{}}Avg.\\ Cost\end{tabular}} & \multicolumn{1}{c}{\begin{tabular}[c]{@{}c@{}}Avg.\\ Time\end{tabular}} & \multicolumn{1}{c}{} & \multicolumn{1}{c}{\begin{tabular}[c]{@{}c@{}}Best\\ Cost\end{tabular}} & \multicolumn{1}{c}{\begin{tabular}[c]{@{}c@{}}Avg.\\ Cost\end{tabular}} & \multicolumn{1}{c}{\begin{tabular}[c]{@{}c@{}}Avg.\\ Time\end{tabular}} & \multicolumn{1}{c}{} & \multicolumn{1}{c}{\begin{tabular}[c]{@{}c@{}}Best\\ Cost\end{tabular}} & \multicolumn{1}{c}{\begin{tabular}[c]{@{}c@{}}Avg.\\ Cost\end{tabular}} & \multicolumn{1}{c}{\begin{tabular}[c]{@{}c@{}}Avg.\\ Time\end{tabular}} \\ \hline
19 & 11686.39\tnote{d} & 11742.78 & 11782.36 & 1117.8 &  & 11696.58 & 11772.45 & 998.8 &  & \textbf{11687.12} & \textbf{11725.38}\tnote{\dag\ddag} & \textbf{676.4} \\
22 & 13068.03\tnote{a} & 13130.75 & 13150.24 & 1126.1 &  & 13123.16 & 13153.59 & 994.0 &  & \textbf{13094.74} & \textbf{13145.70} & \textbf{587.6} \\
23 & 7741.01\tnote{d} & 7794.58 & 7825.57 & 743.1 &  & 7792.16 & 7827.61 & 641.0 &  & \textbf{7752.05} & \textbf{7783.74}\tnote{\dag\ddag} & \textbf{487.5} \\
27 & 8417.62\tnote{c} & 8429.44 & 8459.19 & 1396.9 &  & 8433.13 & 8458.55 & 1232.4 &  & \textbf{8422.36} & \textbf{8441.02}\tnote{\dag\ddag} & \textbf{598.1} \\
32 & 9378.30\tnote{c} & 9445.69 & 9505.94 & 1661.1 &  & 9447.41 & 9489.79 & 1434.6 &  & {\ul \textbf{9348.55}} & \textbf{9404.36}\tnote{\dag\ddag} & \textbf{1045.4} \\
38 & 11194.68\tnote{d} & 11212.19 & 11272.86 & 1048.4 &  & 11245.64 & 11278.79 & 930.4 &  & {\ul \textbf{11192.74}} & \textbf{11227.01}\tnote{\dag\ddag} & \textbf{683.0} \\
46 & 24428.54\tnote{b} & 24607.77 & 24704.02 & 2128.1 &  & 24537.33 & 24694.08 & 1845.3 &  & {\ul \textbf{24404.42}} & \textbf{24580.68}\tnote{\dag\ddag} & \textbf{1345.2} \\
49 & 16181.17\tnote{d} & 16268.83 & 16365.17 & 2404.3 &  & 16197.43 & 16334.44 & 2091.5 &  & {\ul \textbf{16164.00}} & \textbf{16257.90}\tnote{\dag\ddag} & \textbf{1779.7} \\
58 & 23370.42\tnote{d} & 23628.32 & 23781.84 & 1013.9 &  & 23549.77 & 23776.21 & 851.5 &  & \textbf{23396.28} & \textbf{23545.34}\tnote{\dag\ddag} & \textbf{659.3} \\
62 & 22952.06\tnote{b} & 23064.45 & 23195.80 & 1312.8 &  & 23065.38 & 23219.89 & 1204.5 &  & {\ul \textbf{22903.99}} & \textbf{23043.85}\tnote{\dag\ddag} & \textbf{690.7} \\
65 & 13013.89\tnote{b} & 13021.94 & 13074.23 & 1442.5 &  & 13027.21 & 13075.46 & 1110.4 &  & {\ul \textbf{12975.38}} & \textbf{13049.25}\tnote{\dag\ddag} & \textbf{583.6} \\ \hline
\multicolumn{2}{l}{\# of wins} & - & - & - &  & - & - & - &  & \textbf{11} & \textbf{11} & \textbf{11} \\
\multicolumn{2}{l}{APD} &  &  &  &  & -0.10\% & -0.03\% & -14.34\% &  & \textbf{-0.56\%} & \textbf{-0.52\%} & \textbf{-53.24\%} \\ \hline
\end{tabular}
\begin{tablenotes}
\item [\dag]Statistically significant (MS-ILS vs. MineReduce)
\item [\ddag]Statistically significant (MDM-MS-ILS vs. MineReduce)
\item [a]\citet{duhameletal2013}
\item [b]\citet{pennaetal2013b}
\item [c]\citet{kochetov_khmelev2015}
\item [d]\citet{pennaetal2019}
\end{tablenotes}
\end{threeparttable}
\end{center}
\end{table}

Finally, we have compared the results obtained by MineReduce in our experiments to the results reported in the literature for two other state-of-the-art algorithms: HLS, presented by \citet{kochetov_khmelev2015}; and HILS-RVRP, presented by \citet{pennaetal2019}.

The experiments with HLS were run on an Intel\textsuperscript{\textregistered} Core\textsuperscript{TM} i7 2.20 GHz CPU (the specific model was not reported), whereas those with HILS-RVRP were run on an Intel\textsuperscript{\textregistered} Core\textsuperscript{TM} i7-940 2.93 GHz CPU. It must be noticed that in both cases only ten runs were performed per instance, whereas we have performed 20 runs of MineReduce per instance in our experiments.

Since our experiments with MineReduce and those reported with HLS and HILS-RVRP have been performed on different CPU models, a fully precise comparison regarding computational time is not possible. Hence, in order to make this comparison as fair as possible, we consider an approximate scale ratio to compensate for the performance differences between the CPUs. The scale ratios we have adopted were based on the performance ratings from the PassMark CPU benchmarks \citep{passmark2020}. Specifically, as all three algorithms were tested on single threads, we have used the single-thread performance ratings from the PassMark benchmarks.

As the specific CPU model used in the experiments with HLS could not be retrieved, we have assumed it to be the lowest-performance one among all i7 2.20 GHz models (in order to derive approximate lower bounds for the computational time reported). That was the i7-2675QM model. The single-thread ratings for the i7-2675QM 2.20 GHz (assumed for HLS), the i7-940 2.93 GHz (used for HILS-RVRP) and the i7-5500U 2.40 GHz (used for MineReduce) were 1101, 1334 and 1551, respectively. Therefore, the computational times reported for HLS have been multiplied by the ratio 1101/1551, whereas the ratio 1334/1551 has been used for the HILS-RVRP computational times.

Table \ref{tab:SOTA} presents a summary of this comparison (the detailed comparison is presented in \ref{sec:SOTA}). HILS-RVRP and MineReduce have presented the best results regarding both solution quality and computational time. The APD values presented for these algorithms are relative to HLS. Regarding solution quality, a balance is observed for the smallest instances (Set 1), whereas HILS-RVRP has the best results for medium-size instances (Set 2), and MineReduce presents the best results for the largest instances (Sets 3 and 4). Computational times reported for HILS-RVRP and MineReduce are much shorter than those reported for HLS. Between HILS-RVRP and MineReduce, the former presents shorter computational times.

\begin{table}[!ht] \scriptsize
\setlength{\tabcolsep}{5pt}
\begin{center}
\caption{Summary of the comparison of MineReduce to other state-of-the-art algorithms}
\label{tab:SOTA}
\begin{tabular}{llrrrrrrrrrrr}
\hline
 &  & \multicolumn{3}{c}{HLS} & \multicolumn{1}{c}{} & \multicolumn{3}{c}{HILS-RVRP} & \multicolumn{1}{c}{} & \multicolumn{3}{c}{MineReduce} \\ \cline{3-5} \cline{7-9} \cline{11-13} 
 &  & \multicolumn{1}{c}{\begin{tabular}[c]{@{}c@{}}Best\\ Cost\end{tabular}} & \multicolumn{1}{c}{\begin{tabular}[c]{@{}c@{}}Avg.\\ Cost\end{tabular}} & \multicolumn{1}{c}{\begin{tabular}[c]{@{}c@{}}Avg.\\ Time\end{tabular}} & \multicolumn{1}{c}{} & \multicolumn{1}{c}{\begin{tabular}[c]{@{}c@{}}Best\\ Cost\end{tabular}} & \multicolumn{1}{c}{\begin{tabular}[c]{@{}c@{}}Avg.\\ Cost\end{tabular}} & \multicolumn{1}{c}{\begin{tabular}[c]{@{}c@{}}Avg.\\ Time\end{tabular}} & \multicolumn{1}{c}{} & \multicolumn{1}{c}{\begin{tabular}[c]{@{}c@{}}Best\\ Cost\end{tabular}} & \multicolumn{1}{c}{\begin{tabular}[c]{@{}c@{}}Avg.\\ Cost\end{tabular}} & \multicolumn{1}{c}{\begin{tabular}[c]{@{}c@{}}Avg.\\ Time\end{tabular}} \\ \hline
Set 1 & \# of wins & 5 & \textbf{6} & 1 &  & \textbf{13} & \textbf{6} & \textbf{10} &  & 10 & \textbf{6} & 3 \\
 & APD &  &  &  &  & \textbf{-0.09\%} & \textbf{-0.09\%} & \textbf{-119.22\%} &  & -0.08\% & -0.03\% & -89.85\% \\
Set 2 & \# of wins & 7 & 5 & - &  & \textbf{26} & \textbf{21} & \textbf{37} &  & 19 & 11 & - \\
 & APD &  &  &  &  & \textbf{-0.22\%} & \textbf{-0.21\%} & \textbf{-171.14\%} &  & -0.19\% & -0.18\% & -140.31\% \\
Set 3 & \# of wins & 2 & 3 & - &  & \textbf{16} & 7 & \textbf{29} &  & 13 & \textbf{20} & 1 \\
 & APD &  &  &  &  & \textbf{-0.54\%} & -0.32\% & \textbf{-167.21\%} &  & -0.51\% & \textbf{-0.50\%} & -147.24\% \\
Set 4 & \# of wins & 1 & 1 & - &  & 4 & 2 & \textbf{10} &  & \textbf{6} & \textbf{8} & 1 \\
 & APD &  &  &  &  & -0.36\% & -0.22\% & \textbf{-176.96\%} &  & \textbf{-0.53\%} & \textbf{-0.43\%} & -156.42\% \\ \hline
Global & \# of wins & 15 & 15 & 1 &  & \textbf{59} & 36 & \textbf{86} &  & 48 & \textbf{45} & 5 \\
 & APD &  &  &  &  & \textbf{-0.32\%} & -0.23\% & \textbf{-162.46\%} &  & \textbf{-0.32\%} & \textbf{-0.30\%} & -136.74\% \\ \hline
\end{tabular}
\end{center}
\end{table}

\subsection{Behavior analysis} \label{sec:behavior}

To further inspect the behavior of the new MineReduce-based hybrid heuristic, additional experiments were performed using instance 02 from Set 3 of Duhamel et al. These experiments were motivated by those performed in the behavior analysis presented by \citet{maiaetal2018}.

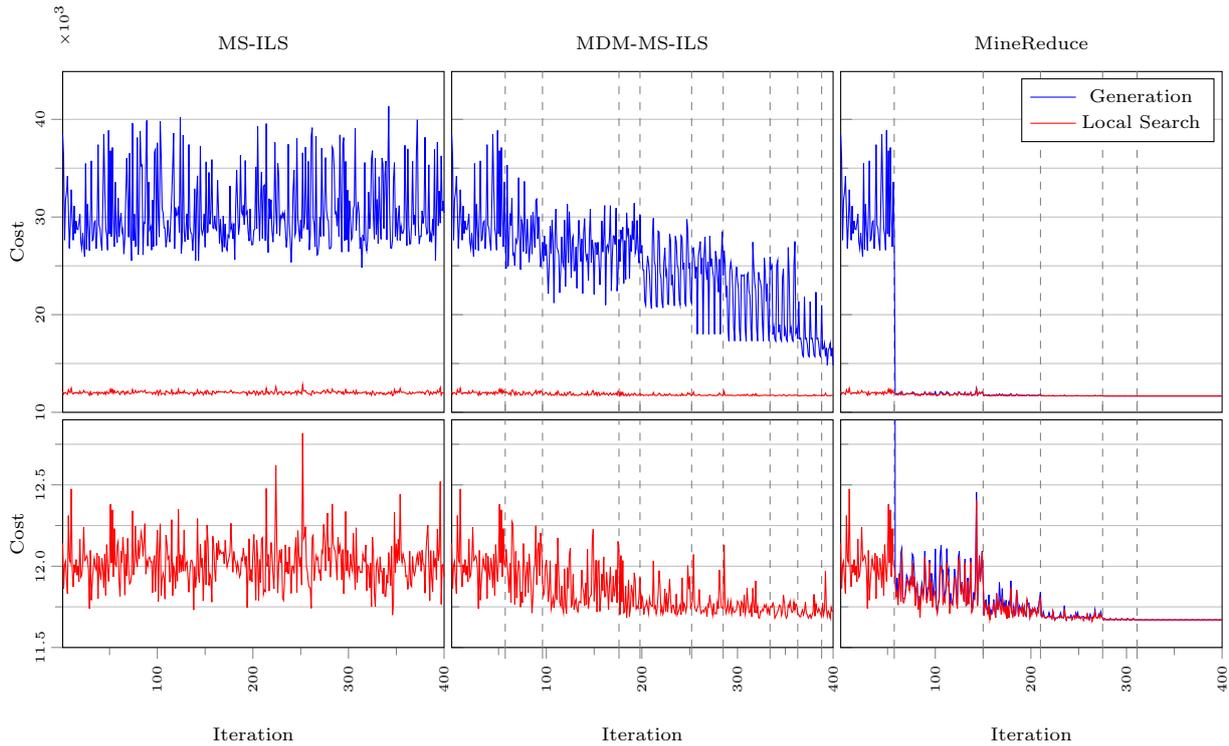
\begin{figure}[!ht]
\centering
\begin{tikzpicture}
	\begin{groupplot}[
			group style={
				group size=3 by 2,
				y descriptions at=edge left,
				x descriptions at=edge bottom,
				horizontal sep=1mm,
				vertical sep=1mm,
				/pgf/number format/.cd,
				1000 sep={},
			},
			height=6.1cm,
			width=6.6cm,
			xlabel={\scriptsize{Iteration}},
			ylabel={\scriptsize{Cost}},
			ylabel near ticks,
			xmin=1,
			xmax=400,
			ymin=10000,
			ymax=44900,
			ymajorgrids=true,
			yminorgrids=true,
			xtick align=outside,
			tickpos=left,
			xticklabel style={rotate=90,anchor=east},
			yticklabel style={rotate=90},
			y tick scale label style={
				at={(0.00,1.20)},anchor=east,inner sep=0pt,
			},
			minor x tick num=1,
			minor y tick num=1,
		]
		\tiny
    
		\nextgroupplot[
			title={\scriptsize{MS-ILS}},
			ytick align=outside,
			xtick=\empty,
			xtick align=inside,
			scaled y ticks=base 10:-3,
			ytick scale label code/.code={$\times 10^3$},
		]
		\addplot+[mark=none] table[x index=0,y index=1] {Duh-02_PRVFHF_s3_m400.txt};
		\addplot+[mark=none] table[x index=0,y index=3] {Duh-02_PRVFHF_s3_m400.txt};
		
		\nextgroupplot[
			title={\scriptsize{MDM-MS-ILS}},
			xtick=\empty,
			xtick align=inside,
			extra x ticks={57,96,176,198,252,285,334,363,388},
			extra x tick style={grid=major,grid style={gray,dashed},xticklabel=\empty},
			scaled ticks=false,
		]
		\addplot+[mark=none] table[x index=0,y index=5] {Duh-02_PRVFHF_s3_m400.txt};
		\addplot+[mark=none] table[x index=0,y index=7] {Duh-02_PRVFHF_s3_m400.txt};
		
		\nextgroupplot[
			title={\scriptsize{MineReduce}},
			xtick=\empty,
			xtick align=inside,
			extra x ticks={57,150,210,275,311},
			extra x tick style={grid=major,grid style={gray,dashed},xticklabel=\empty},
			scaled ticks=false,
		]
		\addplot+[mark=none] table[x index=0,y index=9] {Duh-02_PRVFHF_s3_m400.txt};
		\addlegendentry{\scriptsize{Generation}}
		\addplot+[mark=none] table[x index=0,y index=11] {Duh-02_PRVFHF_s3_m400.txt};
		\addlegendentry{\scriptsize{Local Search}}
    
		\nextgroupplot[
			height=4.6cm,
			ymin=11500,
			ymax=12900,
			ytick align=outside,
			xlabel shift={8.5pt},
			scaled y ticks=base 10:-3,
			ytick scale label code/.code={$ $},
			y tick label style={
        /pgf/number format/.cd,
            fixed,
            fixed zerofill,
            precision=1,
        /tikz/.cd
			},
		]
		\addplot+[mark=none] table[x index=0,y index=1] {Duh-02_PRVFHF_s3_m400.txt};
		\addplot+[mark=none] table[x index=0,y index=3] {Duh-02_PRVFHF_s3_m400.txt};
		
		\nextgroupplot[
			height=4.6cm,
			ymin=11500,
			ymax=12900,
			extra x ticks={57,96,176,198,252,285,334,363,388},
			extra x tick style={grid=major,grid style={gray,dashed},tick label style={teal,rotate=-35}},
			extra x tick labels={},
			scaled ticks=false,
		]
		\addplot+[mark=none] table[x index=0,y index=5] {Duh-02_PRVFHF_s3_m400.txt};
		\addplot+[mark=none] table[x index=0,y index=7] {Duh-02_PRVFHF_s3_m400.txt};
		
		\nextgroupplot[
			height=4.6cm,
			ymin=11500,
			ymax=12900,
			extra x ticks={57,150,210,275,311},
			extra x tick style={grid=major,grid style={gray,dashed},tick label style={teal,rotate=-35}},
			extra x tick labels={},
			scaled ticks=false,
		]
		\addplot+[mark=none] table[x index=0,y index=9] {Duh-02_PRVFHF_s3_m400.txt};
		\addplot+[mark=none] table[x index=0,y index=11] {Duh-02_PRVFHF_s3_m400.txt};
	\end{groupplot}
\end{tikzpicture}
\caption{Cost vs.~iteration charts illustrating the behavior of MS-ILS, MDM-MS-ILS and MineReduce for instance 02 from Set 3 of Duhamel et al. over 400 iterations}
\label{fig:costChartA}
\end{figure}

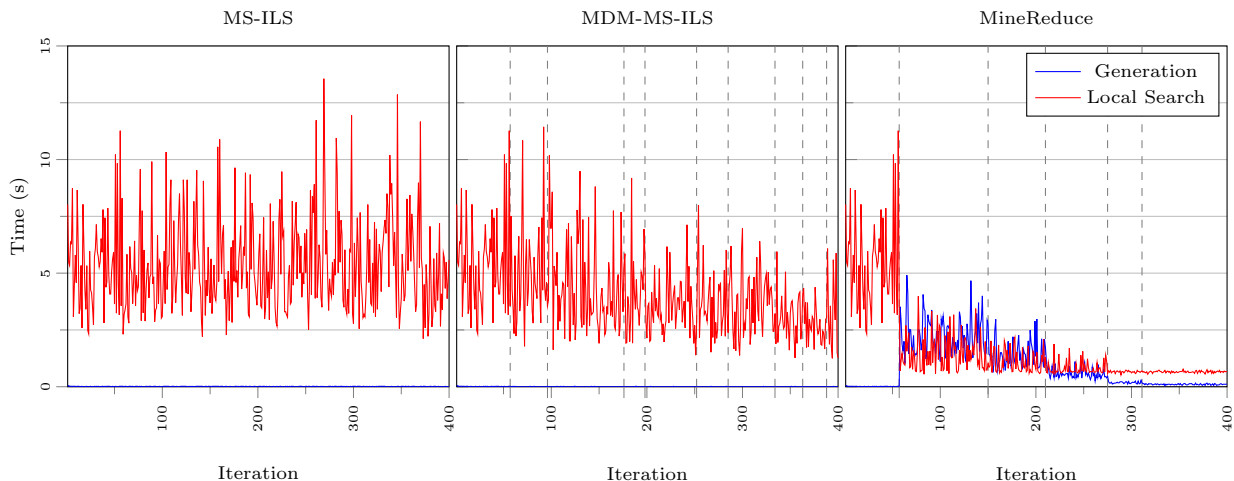
\begin{figure}[!ht]
\centering
\begin{tikzpicture}
	\begin{groupplot}[
			group style={
				group size=3 by 1,
				y descriptions at=edge left,
				x descriptions at=edge bottom,
				horizontal sep=1mm,
				vertical sep=1mm,
				/pgf/number format/.cd,
				1000 sep={},
			},
			height=6.1cm,
			width=6.6cm,
			xlabel={\scriptsize{Iteration}},
			ylabel={\scriptsize{Time (s)}},
			ylabel near ticks,
			xmin=1,
			xmax=400,
			ymin=0,
			ymax=15,
			ymajorgrids=true,
			yminorgrids=true,
			xtick align=outside,
			tickpos=left,
			xticklabel style={rotate=90,anchor=east},
			yticklabel style={rotate=90},
			minor x tick num=1,
			minor y tick num=1,
		]
		\tiny
    
		\nextgroupplot[
			title={\scriptsize{MS-ILS}},
			ytick align=outside,
			xlabel shift={8.5pt},
		]
		\addplot+[mark=none] table[x index=0,y index=2] {Duh-02_PRVFHF_s3_m400.txt};
		\addplot+[mark=none] table[x index=0,y index=4] {Duh-02_PRVFHF_s3_m400.txt};
		
		\nextgroupplot[
			title={\scriptsize{MDM-MS-ILS}},
			extra x ticks={57,96,176,198,252,285,334,363,388},
			extra x tick style={grid=major,grid style={gray,dashed},tick label style={teal,rotate=-35}},
			extra x tick labels={},
		]
		\addplot+[mark=none] table[x index=0,y index=6] {Duh-02_PRVFHF_s3_m400.txt};
		\addplot+[mark=none] table[x index=0,y index=8] {Duh-02_PRVFHF_s3_m400.txt};
		
		\nextgroupplot[
			title={\scriptsize{MineReduce}},
			extra x ticks={57,150,210,275,311},
			extra x tick style={grid=major,grid style={gray,dashed},tick label style={teal,rotate=-35}},
			extra x tick labels={},
		]
		\addplot+[mark=none] table[x index=0,y index=10] {Duh-02_PRVFHF_s3_m400.txt};
		\addlegendentry{\scriptsize{Generation}}
		\addplot+[mark=none] table[x index=0,y index=12] {Duh-02_PRVFHF_s3_m400.txt};
		\addlegendentry{\scriptsize{Local Search}}
	\end{groupplot}
\end{tikzpicture}
\caption{Time vs.~iteration charts illustrating the behavior of MS-ILS, MDM-MS-ILS and MineReduce for instance 02 from Set 3 of Duhamel et al. over 400 iterations}
\label{fig:timeChartA}
\end{figure}

In the first of these experiments, the heuristics have been run with 400 iterations. Fig. \ref{fig:costChartA} provides an evaluation of the solution costs obtained throughout the experiment. The charts in the first row show, for each heuristic, the solution costs obtained per iteration in the generation and local search phases, whereas the second row provides enlarged views focusing on the bottom part. In Fig. \ref{fig:timeChartA}, the charts exhibit the computational time spent in the generation and local search phases per iteration. The dashed vertical lines indicate the iterations preceding a data mining method run.

In Fig. \ref{fig:costChartA}, the charts in the first row show that the reduction in the costs of the solutions generated by MineReduce after data mining is much more expressive than that observed for MDM-MS-ILS. The enlarged views in the second row show that the costs of generated solutions fall to a level even lower than that of the costs of solutions discovered through the local search in previous iterations, and the local search finds even better solutions after each run of the data mining procedure.

On the other hand, as shown in the last chart of Fig. \ref{fig:timeChartA}, the computational time spent in the generation phase -- which is close to zero in the first iterations -- increases after the first run of the data mining procedure. This increase is due to the execution of the PSR process, which includes a local search on a reduced version of the problem instance. However, the increase in time spent in the generation phase is compensated by a reduction, quite expressive, in time spent in the local search phase. The amount of time spent in the generation and local search phases together per iteration is significantly reduced after data mining.

The overall reduction in computational time can be explained by the fact that MineReduce shrinks the problem instance and, consequently, the search space. Therefore, the first local search (embedded in the hybrid generation phase, over the reduced instance) performs a much smaller number of movement evaluations, thus it is much faster. Afterwards, the second local search (actual local search phase, over the original instance) starts from a high-quality solution, so it converges faster as well.

In MineReduce, as the charts show, the benefits -- reduction of solution costs and computational time -- are intensified after each execution of the data mining procedure, reaching very low levels in the last iterations, which explains the superiority demonstrated by the results of the experiments presented in Section \ref{subsec:resultsHFFVRP}. In this experiment, a new best solution to instance 02 of Duhamel et al., with a cost of 11660.12, was discovered by MineReduce at the 264\textsuperscript{th} iteration.

\begin{figure}[!ht]
\centering
\begin{tikzpicture}
	\begin{groupplot}[
			group style={
				group size=1 by 1,
				y descriptions at=edge left,
				x descriptions at=edge bottom,
				horizontal sep=1mm,
				vertical sep=1mm,
				/pgf/number format/.cd,
				1000 sep={},
			},
			height=6.1cm,
			width=11.1cm,
			xlabel=Time to reach the target (s),
			ylabel=Probability,
			ylabel near ticks,
			xmin=0,
			xmax=600,
			ymin=0,
			ymax=1,
			ymajorgrids=true,
			yminorgrids=true,
			xmajorgrids=true,
			xminorgrids=true,
			xtick align=outside,
			ytick align=outside,
			tickpos=left,
			minor x tick num=1,
			minor y tick num=1,
			grid style={dashed},
			scaled x ticks=false,
			legend pos=south east,
		]
		\scriptsize
    
		\nextgroupplot
		\addplot+[only marks,mark=star,mark options={red}] table[x index=1,y index=0] {Duh-02_PRVFHF_TTT.txt};
		\addlegendentry{MS-ILS}
		\addplot+[only marks,mark=+,mark options={blue}] table[x index=2,y index=0] {Duh-02_PRVFHF_TTT.txt};
		\addlegendentry{MDM-MS-ILS}
		\addplot+[only marks,mark=Mercedes star flipped,mark options={green}] table[x index=3,y index=0] {Duh-02_PRVFHF_TTT.txt};
		\addlegendentry{MineReduce}
	\end{groupplot}
\end{tikzpicture}
\caption{TTT plots comparing all heuristics for instance 02 from Set 3 of \citet{duhameletal2010}}
\label{fig:TTTDM}
\end{figure}
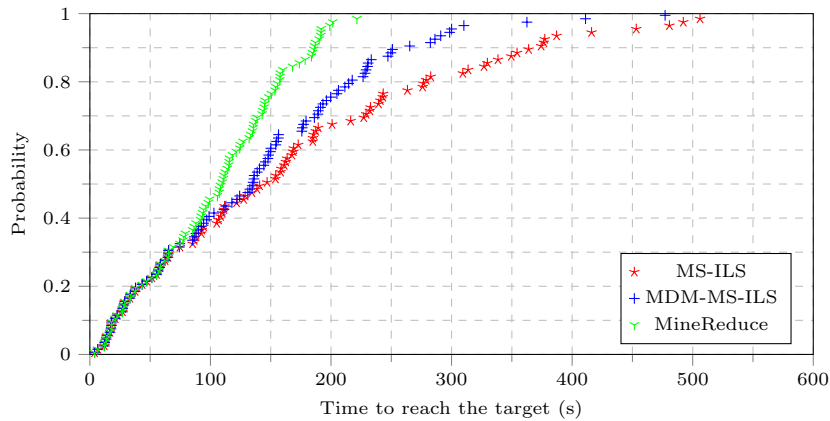

The second experiment focused on the generation and evaluation of time-to-target (TTT) plots~\citep{aiexetal2007}. A TTT plot displays, on the ordinate axis, the probability that an algorithm will discover a solution at least as good as a given target cost within a given running time, which is shown on the abscissa axis. In this experiment, each heuristic was run 100 times, with 100 distinct random seeds, targeting a solution with a cost lower than or equal to 11780. The chart obtained is shown in Fig. \ref{fig:TTTDM}.

The chart shows that MineReduce outperforms the other heuristics. It is possible to observe that the probability that the target will be reached within 200 seconds, for example, is approximately 97\% for MineReduce, 76\% for MDM-MS-ILS, and 67\% for MS-ILS.

The analysis of the charts presented in Figs. \ref{fig:costChartA}, \ref{fig:timeChartA} and \ref{fig:TTTDM} clearly illustrates the influence of the MineReduce approach in the behavior of the heuristics. As expected, this behavior causes a performance improvement, regarding both the quality of solutions found and the computational time spent.

\section{Conclusion} \label{sec:conclusion}

Previous work that explored data science in combinatorial optimization produced highly significant results by applying patterns (found by data mining procedures) to guide the construction of initial solutions.

This work presents an approach that uses mined patterns to perform problem size reduction. The MineReduce approach was applied to extend a previous and state-of-the-art heuristic for the HFVRP \citep{pennaetal2013a}. The new hybrid heuristic obtained, named MineReduce-MS-ILS, produced significantly better results in terms of both solution quality and computational time when compared to the original heuristic and a previous hybrid version with data mining.

We carried out experiments on the 96 HFVRP benchmark instances from the sets of \citet{duhameletal2010} (four reserved for parameter tuning). The results attained show that the proposed approach is very promising, as the MineReduce-based heuristic reached the best average solution costs and computational times for most instances. Moreover, it obtained new upper bounds for 22 instances.

The proposed heuristic presented better performances than the MS-ILS heuristic and a previous hybrid version with data mining (MDM-MS-ILS). MineReduce attained better average costs for 83\% of the instances (65\% with statistical significance) in comparison to the MS-ILS and better average costs for 76\% of the instances (64\% with statistical significance) in comparison to the MDM-MS-ILS. If small instances (Set~1) -- for which the three heuristics have presented similar performance -- are excluded from the comparison, than the superiority of MineReduce is clearer revealed: better average costs for 91\% of the instances (76\% with statistical significance) in comparison to the MS-ILS and better average costs for 86\% of the instances (74\% with statistical significance) in comparison to the MDM-MS-ILS.

Furthermore, we have compared the results obtained by MineReduce in our experiments to those reported in the literature for two other state-of-the-art algorithms: HLS, presented by \citet{kochetov_khmelev2015}; and HILS-RVRP, presented by \citet{pennaetal2019}. This comparison showed that MineReduce is very competitive, especially for large instances.

The reported results are evidence that heuristics based on MineReduce can generate initial solutions of better quality. Therefore, a significant improvement in the quality of solutions obtained throughout the local search phase is observed as well. Additionally, a consistent reduction of the convergence time of the local search phase is also noticed. Therefore, the proposed MineReduce approach shall be further studied and explored in future work, including its application to other heuristics and other optimization problems.

\appendix
\section{Detailed comparison of MineReduce to other state-of-the-art algorithms} \label{sec:SOTA}

This appendix presents a detailed comparison of the results obtained by MineReduce in our experiments to the results reported in the literature for other state-of-the-art algorithms: HLS \citep{kochetov_khmelev2015} and HILS-RVRP \citep{pennaetal2019}.

Tables \ref{tab:SOTA1}, \ref{tab:SOTA2}, \ref{tab:SOTA3} and \ref{tab:SOTA4} show the comparisons on instances of Sets 1, 2, 3 and 4 of \citet{duhameletal2010}, respectively.

\begin{table}[!ht] \scriptsize
\setlength{\tabcolsep}{5pt}
\begin{center}
\caption{Detailed comparison of MineReduce to other state-of-the-art algorithms -- Set 1 of \citet{duhameletal2010}}
\label{tab:SOTA1}
\begin{tabular}{crrrrrrrrrrr}
\hline
 & \multicolumn{3}{c}{HLS} & \multicolumn{1}{c}{} & \multicolumn{3}{c}{HILS-RVRP} & \multicolumn{1}{c}{} & \multicolumn{3}{c}{MineReduce} \\ \cline{2-4} \cline{6-8} \cline{10-12} 
I & \multicolumn{1}{c}{\begin{tabular}[c]{@{}c@{}}Best\\ Cost\end{tabular}} & \multicolumn{1}{c}{\begin{tabular}[c]{@{}c@{}}Avg.\\ Cost\end{tabular}} & \multicolumn{1}{c}{\begin{tabular}[c]{@{}c@{}}Avg.\\ Time\end{tabular}} & \multicolumn{1}{c}{} & \multicolumn{1}{c}{\begin{tabular}[c]{@{}c@{}}Best\\ Cost\end{tabular}} & \multicolumn{1}{c}{\begin{tabular}[c]{@{}c@{}}Avg.\\ Cost\end{tabular}} & \multicolumn{1}{c}{\begin{tabular}[c]{@{}c@{}}Avg.\\ Time\end{tabular}} & \multicolumn{1}{c}{} & \multicolumn{1}{c}{\begin{tabular}[c]{@{}c@{}}Best\\ Cost\end{tabular}} & \multicolumn{1}{c}{\begin{tabular}[c]{@{}c@{}}Avg.\\ Cost\end{tabular}} & \multicolumn{1}{c}{\begin{tabular}[c]{@{}c@{}}Avg.\\ Time\end{tabular}} \\ \hline
08 & 4596.52 & 4597.65 & 180.3 &  & \textbf{4591.75} & 4595.65 & \textbf{10.3} &  & \textbf{4591.75} & \textbf{4594.20} & 30.2 \\
10 & 2108.10 & 2108.32 & 131.3 &  & \textbf{2107.55} & 2107.83 & \textbf{10.4} &  & \textbf{2107.55} & \textbf{2107.55} & 29.5 \\
11 & \textbf{3367.41} & \textbf{3372.57} & 305.2 &  & \textbf{3367.41} & 3373.77 & \textbf{19.0} &  & \textbf{3367.41} & 3376.12 & 49.0 \\
36 & 5709.31 & 5738.34 & 254.1 &  & \textbf{5684.61} & \textbf{5700.14} & \textbf{25.6} &  & 5684.62 & 5702.82 & 70.5 \\
39 & 2926.59 & \textbf{2928.99} & 143.4 &  & 2921.40 & 2932.75 & \textbf{16.8} &  & \textbf{2920.93} & 2933.75 & 47.1 \\
43 & 8737.13 & 8749.47 & 352.1 &  & \textbf{8737.02} & \textbf{8746.38} & 64.8 &  & 8739.36 & 8756.97 & \textbf{55.7} \\
52 & 4035.59 & 4035.59 & 43.3 &  & \textbf{4027.27} & 4030.44 & \textbf{13.8} &  & 4029.42 & \textbf{4030.35} & 15.6 \\
55 & 10256.16 & 10264.37 & 32.7 &  & \textbf{10244.34} & \textbf{10250.98} & \textbf{11.4} &  & \textbf{10244.34} & 10313.29 & 14.0 \\
70 & 6689.61 & 6729.87 & 142.7 &  & \textbf{6684.56} & \textbf{6694.43} & \textbf{13.5} &  & 6685.24 & 6695.22 & 29.8 \\
75 & \textbf{452.85} & \textbf{452.85} & \textbf{0.7} &  & \textbf{452.85} & \textbf{452.85} & 1.0 &  & \textbf{452.85} & \textbf{452.85} & 3.2 \\
82 & 4769.35 & 4772.58 & 127.8 &  & \textbf{4766.74} & 4771.33 & 31.2 &  & \textbf{4766.74} & \textbf{4770.34} & \textbf{30.0} \\
92 & \textbf{564.39} & \textbf{564.39} & 20.6 &  & \textbf{564.39} & 564.65 & \textbf{3.9} &  & \textbf{564.39} & \textbf{564.39} & 9.1 \\
93 & \textbf{1036.99} & \textbf{1036.99} & 13.5 &  & \textbf{1036.99} & 1038.34 & \textbf{6.0} &  & \textbf{1036.99} & 1038.38 & 8.6 \\
94 & \textbf{1378.25} & \textbf{1378.25} & 31.9 &  & \textbf{1378.25} & \textbf{1378.25} & 27.0 &  & \textbf{1378.25} & 1378.34 & \textbf{17.1} \\ \hline
\multicolumn{1}{l}{\# of wins} & 5 & \textbf{6} & 1 &  & \textbf{13} & \textbf{6} & \textbf{10} &  & 10 & \textbf{6} & 3 \\
\multicolumn{1}{l}{APD} &  &  &  &  & \textbf{-0.09\%} & \textbf{-0.09\%} & \textbf{-119.22\%} &  & -0.08\% & -0.03\% & -89.85\% \\ \hline
\end{tabular}
\end{center}
\end{table}

\begin{table}[!ht] \scriptsize
\setlength{\tabcolsep}{5pt}
\begin{center}
\caption{Detailed comparison of MineReduce to other state-of-the-art algorithms -- Set 2 of \citet{duhameletal2010}}
\label{tab:SOTA2}
\begin{tabular}{crrrrrrrrrrr}
\hline
 & \multicolumn{3}{c}{HLS} & \multicolumn{1}{c}{} & \multicolumn{3}{c}{HILS-RVRP} & \multicolumn{1}{c}{} & \multicolumn{3}{c}{MineReduce} \\ \cline{2-4} \cline{6-8} \cline{10-12} 
I & \multicolumn{1}{c}{\begin{tabular}[c]{@{}c@{}}Best\\ Cost\end{tabular}} & \multicolumn{1}{c}{\begin{tabular}[c]{@{}c@{}}Avg.\\ Cost\end{tabular}} & \multicolumn{1}{c}{\begin{tabular}[c]{@{}c@{}}Avg.\\ Time\end{tabular}} & \multicolumn{1}{c}{} & \multicolumn{1}{c}{\begin{tabular}[c]{@{}c@{}}Best\\ Cost\end{tabular}} & \multicolumn{1}{c}{\begin{tabular}[c]{@{}c@{}}Avg.\\ Cost\end{tabular}} & \multicolumn{1}{c}{\begin{tabular}[c]{@{}c@{}}Avg.\\ Time\end{tabular}} & \multicolumn{1}{c}{} & \multicolumn{1}{c}{\begin{tabular}[c]{@{}c@{}}Best\\ Cost\end{tabular}} & \multicolumn{1}{c}{\begin{tabular}[c]{@{}c@{}}Avg.\\ Cost\end{tabular}} & \multicolumn{1}{c}{\begin{tabular}[c]{@{}c@{}}Avg.\\ Time\end{tabular}} \\ \hline
05 & 10896.35 & 10937.73 & 637.5 &  & \textbf{10876.48} & \textbf{10897.93} & \textbf{22.9} &  & \textbf{10876.48} & 10925.52 & 76.4 \\
06 & 11760.08 & 11783.45 & 1031.4 &  & \textbf{11688.64} & \textbf{11734.52} & \textbf{37.5} &  & 11696.83 & 11761.32 & 125.0 \\
07 & \textbf{8074.64} & 8135.00 & 398.9 &  & 8089.46 & 8144.80 & \textbf{23.6} &  & 8076.53 & \textbf{8129.35} & 68.6 \\
12 & \textbf{3543.99} & \textbf{3545.82} & 448.6 &  & \textbf{3543.99} & 3547.92 & \textbf{49.5} &  & \textbf{3543.99} & 3547.65 & 82.9 \\
13 & 6709.28 & 6728.61 & 606.2 &  & \textbf{6696.43} & \textbf{6703.23} & \textbf{43.8} &  & 6697.58 & 6706.94 & 102.8 \\
16 & \textbf{4156.97} & \textbf{4160.43} & 767.4 &  & \textbf{4156.97} & 4164.03 & \textbf{59.1} &  & \textbf{4156.97} & 4163.04 & 126.1 \\
17 & 5381.19 & 5403.10 & 311.6 &  & \textbf{5362.83} & \textbf{5367.76} & \textbf{36.2} &  & 5365.94 & 5381.80 & 61.3 \\
2A & 7820.37 & 7862.82 & 851.8 &  & \textbf{7793.16} & \textbf{7796.54} & \textbf{37.3} &  & \textbf{7793.16} & 7809.81 & 113.1 \\
2B & 8577.50 & 8601.68 & 609.8 &  & 8462.56 & 8499.95 & \textbf{46.8} &  & \textbf{8453.35} & \textbf{8477.33} & 137.0 \\
21 & 5160.03 & 5175.71 & 592.7 &  & \textbf{5139.84} & 5166.11 & \textbf{40.6} &  & \textbf{5139.84} & \textbf{5162.74} & 110.3 \\
25 & 7209.61 & \textbf{7218.87} & 1730.6 &  & \textbf{7209.29} & 7230.50 & \textbf{106.5} &  & 7209.50 & 7231.23 & 281.3 \\
26 & \textbf{6393.47} & \textbf{6433.33} & 760.3 &  & 6433.21 & 6461.05 & \textbf{128.2} &  & 6438.69 & 6458.32 & 275.7 \\
28 & 5538.45 & 5550.86 & 1107.4 &  & 5530.55 & 5542.80 & \textbf{87.6} &  & \textbf{5529.05} & \textbf{5537.72} & 171.4 \\
30 & 6329.09 & 6361.97 & 390.4 &  & \textbf{6315.70} & \textbf{6342.42} & \textbf{71.3} &  & 6331.40 & 6350.05 & 117.9 \\
31 & 4105.67 & 4130.97 & 1456.6 &  & \textbf{4091.52} & \textbf{4112.64} & \textbf{87.9} &  & \textbf{4091.52} & 4113.60 & 218.8 \\
34 & 5784.25 & 5799.40 & 1110.9 &  & \textbf{5747.25} & \textbf{5785.59} & \textbf{56.4} &  & 5765.08 & 5792.18 & 185.6 \\
40 & 11156.86 & 11184.18 & 994.5 &  & 11118.57 & 11171.17 & \textbf{77.9} &  & \textbf{11111.89} & \textbf{11140.70} & 162.0 \\
41 & 7606.16 & 7643.93 & 1359.4 &  & 7597.27 & 7672.27 & \textbf{58.6} &  & \textbf{7573.24} & \textbf{7637.37} & 204.9 \\
47 & 16291.49 & 16332.46 & 424.5 &  & \textbf{16156.12} & \textbf{16247.77} & \textbf{35.4} &  & \textbf{16156.12} & 16263.38 & 83.9 \\
48 & 21316.55 & 21444.07 & 470.6 &  & \textbf{21309.94} & \textbf{21391.58} & \textbf{39.3} &  & 21329.71 & 21413.41 & 99.0 \\
51 & - & - & - &  & \textbf{7721.47} & \textbf{7787.85} & \textbf{50.3} &  & 7780.04 & 7804.28 & 117.0 \\
53 & 6483.51 & 6497.73 & 470.6 &  & \textbf{6434.83} & 6454.77 & \textbf{31.0} &  & \textbf{6434.83} & \textbf{6448.50} & 80.4 \\
60 & 17073.80 & 17106.54 & 895.8 &  & \textbf{17036.59} & \textbf{17055.35} & \textbf{63.1} &  & 17045.33 & 17082.39 & 125.2 \\
61 & 7308.84 & 7320.97 & 396.8 &  & \textbf{7292.03} & \textbf{7302.40} & \textbf{32.2} &  & \textbf{7292.03} & 7305.89 & 96.7 \\
66 & 12790.56 & 12862.79 & 1404.1 &  & 12783.94 & 12922.52 & \textbf{97.8} &  & \textbf{12772.07} & \textbf{12839.71} & 295.9 \\
68 & - & - & - &  & 8970.63 & 9123.03 & \textbf{58.4} &  & \textbf{8919.16} & \textbf{8992.62} & 153.9 \\
73 & \textbf{10195.13} & 10215.26 & 1161.3 &  & 10195.33 & \textbf{10195.36} & \textbf{63.3} &  & 10203.84 & 10209.61 & 136.7 \\
74 & 11598.92 & 11634.25 & 885.2 &  & \textbf{11586.58} & \textbf{11591.23} & \textbf{70.9} &  & 11586.87 & 11599.16 & 150.6 \\
79 & 7266.75 & 7310.23 & 1197.5 &  & \textbf{7259.54} & \textbf{7289.26} & \textbf{105.2} &  & 7262.02 & 7290.96 & 234.8 \\
81 & \textbf{10675.92} & \textbf{10690.71} & 387.6 &  & 10686.31 & 10700.27 & \textbf{49.9} &  & 10693.70 & 10699.79 & 95.1 \\
83 & 10041.06 & 10050.45 & 950.5 &  & 10020.07 & 10048.17 & \textbf{62.3} &  & \textbf{10019.15} & \textbf{10046.97} & 124.7 \\
84 & 7228.38 & 7244.86 & 480.6 &  & \textbf{7227.88} & \textbf{7237.93} & \textbf{46.8} &  & \textbf{7227.88} & 7267.13 & 72.4 \\
85 & 8812.03 & 8842.44 & 1602.2 &  & \textbf{8773.08} & \textbf{8818.55} & \textbf{78.4} &  & 8827.98 & 8857.80 & 171.6 \\
87 & \textbf{3753.87} & 3757.12 & 362.7 &  & \textbf{3753.87} & 3756.97 & \textbf{27.0} &  & \textbf{3753.87} & \textbf{3755.06} & 66.2 \\
88 & 12406.93 & 12452.74 & 877.4 &  & \textbf{12388.23} & \textbf{12405.80} & \textbf{40.0} &  & 12402.85 & 12447.06 & 118.6 \\
89 & 7099.68 & 7120.97 & 944.8 &  & \textbf{7086.36} & \textbf{7102.98} & \textbf{65.8} &  & 7095.33 & 7110.97 & 140.0 \\
90 & 2350.68 & 2358.22 & 322.3 &  & \textbf{2346.43} & \textbf{2356.31} & \textbf{41.1} &  & \textbf{2346.43} & 2357.59 & 54.8 \\ \hline
\multicolumn{1}{l}{\# of wins} & 7 & 5 & - &  & \textbf{26} & \textbf{21} & \textbf{37} &  & 19 & 11 & - \\
\multicolumn{1}{l}{APD} &  &  &  &  & \textbf{-0.22\%} & \textbf{-0.21\%} & \textbf{-171.14\%} &  & -0.19\% & -0.18\% & -140.31\% \\ \hline
\end{tabular}
\end{center}
\end{table}

\begin{table}[!ht] \scriptsize
\setlength{\tabcolsep}{5pt}
\begin{center}
\caption{Detailed comparison of MineReduce to other state-of-the-art algorithms -- Set 3 of \citet{duhameletal2010}}
\label{tab:SOTA3}
\begin{tabular}{crrrrrrrrrrr}
\hline
 & \multicolumn{3}{c}{HLS} & \multicolumn{1}{c}{} & \multicolumn{3}{c}{HILS-RVRP} & \multicolumn{1}{c}{} & \multicolumn{3}{c}{MineReduce} \\ \cline{2-4} \cline{6-8} \cline{10-12} 
I & \multicolumn{1}{c}{\begin{tabular}[c]{@{}c@{}}Best\\ Cost\end{tabular}} & \multicolumn{1}{c}{\begin{tabular}[c]{@{}c@{}}Avg.\\ Cost\end{tabular}} & \multicolumn{1}{c}{\begin{tabular}[c]{@{}c@{}}Avg.\\ Time\end{tabular}} & \multicolumn{1}{c}{} & \multicolumn{1}{c}{\begin{tabular}[c]{@{}c@{}}Best\\ Cost\end{tabular}} & \multicolumn{1}{c}{\begin{tabular}[c]{@{}c@{}}Avg.\\ Cost\end{tabular}} & \multicolumn{1}{c}{\begin{tabular}[c]{@{}c@{}}Avg.\\ Time\end{tabular}} & \multicolumn{1}{c}{} & \multicolumn{1}{c}{\begin{tabular}[c]{@{}c@{}}Best\\ Cost\end{tabular}} & \multicolumn{1}{c}{\begin{tabular}[c]{@{}c@{}}Avg.\\ Cost\end{tabular}} & \multicolumn{1}{c}{\begin{tabular}[c]{@{}c@{}}Avg.\\ Time\end{tabular}} \\ \hline
04 & 10861.29 & 10892.04 & 3007.7 &  & 10748.17 & 10775.93 & \textbf{147.6} &  & \textbf{10725.96} & \textbf{10772.83} & 329.2 \\
09 & 7671.00 & 7700.14 & 2219.0 &  & \textbf{7603.38} & \textbf{7630.55} & \textbf{200.0} &  & 7618.34 & 7652.09 & 225.6 \\
14 & 5680.64 & 5705.25 & 2227.6 &  & 5657.62 & 5697.17 & \textbf{307.3} &  & \textbf{5654.49} & \textbf{5671.42} & 449.5 \\
15 & 8255.95 & 8273.64 & 4050.5 &  & \textbf{8220.64} & 8285.79 & \textbf{136.2} &  & 8229.06 & \textbf{8266.88} & 412.0 \\
24 & 9145.18 & 9178.51 & 2383.0 &  & \textbf{9119.92} & 9189.22 & \textbf{140.9} &  & 9128.64 & \textbf{9169.28} & 373.5 \\
29 & 9151.41 & 9182.21 & 1566.7 &  & 9142.86 & \textbf{9149.12} & \textbf{199.7} &  & \textbf{9136.41} & 9151.44 & 275.4 \\
33 & 9419.00 & 9452.46 & 4322.4 &  & \textbf{9410.99} & 9471.26 & \textbf{296.7} &  & 9411.12 & \textbf{9440.74} & 505.4 \\
35 & 9605.62 & 9640.89 & 2324.1 &  & \textbf{9555.92} & \textbf{9585.91} & \textbf{123.9} &  & 9565.36 & 9617.15 & 241.2 \\
37 & 6894.98 & 6915.59 & 1751.9 &  & \textbf{6850.77} & 6875.28 & \textbf{211.0} &  & 6854.99 & \textbf{6869.01} & 345.9 \\
42 & 10940.03 & 11012.21 & 3913.5 &  & \textbf{10817.90} & 10995.75 & \textbf{211.9} &  & 10842.11 & \textbf{10918.29} & 767.3 \\
44 & 12549.34 & 12604.56 & 2032.3 &  & \textbf{12191.48} & 12314.24 & \textbf{137.0} &  & 12214.60 & \textbf{12308.14} & 331.0 \\
45 & 10664.81 & 10719.12 & 1787.4 &  & \textbf{10476.25} & 10614.48 & \textbf{126.5} &  & 10497.86 & \textbf{10540.84} & 343.1 \\
50 & 12466.27 & 12482.40 & 3895.7 &  & 12370.94 & 12430.18 & \textbf{322.0} &  & \textbf{12333.66} & \textbf{12394.94} & 836.6 \\
54 & 10433.38 & 10477.30 & 2399.3 &  & 10351.97 & 10435.58 & \textbf{174.8} &  & \textbf{10342.62} & \textbf{10408.42} & 377.3 \\
56 & 31236.61 & 31286.03 & 1600.0 &  & 31030.19 & 31144.98 & \textbf{223.7} &  & \textbf{31020.27} & \textbf{31122.92} & 280.9 \\
57 & \textbf{43378.37} & \textbf{43581.83} & 2046.5 &  & 44781.64 & 44899.36 & \textbf{215.6} &  & 44771.71 & 44854.99 & 288.1 \\
59 & \textbf{14299.28} & \textbf{14324.68} & 3836.8 &  & 14304.46 & 14357.81 & \textbf{268.5} &  & 14310.58 & 14352.36 & 511.9 \\
63 & 20154.56 & 20262.46 & 2616.6 &  & 20022.94 & 20281.49 & \textbf{183.3} &  & \textbf{19946.64} & \textbf{20047.97} & 347.2 \\
64 & 17157.37 & 17180.65 & 1408.4 &  & \textbf{17135.16} & \textbf{17157.79} & \textbf{91.2} &  & \textbf{17135.16} & 17162.28 & 222.0 \\
67 & 10971.29 & 11024.31 & 2517.9 &  & 10884.91 & 10945.00 & \textbf{236.8} &  & \textbf{10861.53} & \textbf{10935.95} & 361.8 \\
69 & 9222.25 & 9270.15 & 1512.0 &  & 9147.54 & 9190.46 & \textbf{101.4} &  & \textbf{9143.84} & \textbf{9190.45} & 204.8 \\
71 & 9976.24 & 9988.03 & 2425.6 &  & \textbf{9834.40} & 9915.73 & \textbf{93.1} &  & 9847.58 & \textbf{9894.85} & 253.6 \\
72 & 5950.67 & 5977.67 & 2588.9 &  & 5903.81 & 5949.29 & \textbf{193.7} &  & \textbf{5895.89} & \textbf{5929.19} & 469.2 \\
76 & 12007.57 & \textbf{12036.84} & 3027.6 &  & \textbf{11994.40} & 12040.78 & \textbf{119.0} &  & 12009.05 & 12040.47 & 283.4 \\
77 & 6943.61 & 7036.46 & 2734.4 &  & \textbf{6916.01} & 6974.86 & \textbf{233.8} &  & 6920.92 & \textbf{6959.31} & 639.0 \\
78 & 7101.20 & 7162.22 & 3014.8 &  & \textbf{7053.62} & \textbf{7122.27} & \textbf{450.7} &  & 7062.22 & 7123.49 & 522.9 \\
80 & 6833.02 & 6844.95 & 2033.1 &  & \textbf{6819.71} & 6843.88 & \textbf{182.1} &  & 6826.59 & \textbf{6833.72} & 296.6 \\
86 & 9056.31 & 9085.76 & 1783.2 &  & 9027.84 & \textbf{9048.94} & \textbf{216.8} &  & \textbf{9024.02} & 9050.11 & 236.1 \\
91 & 6374.01 & 6398.88 & 3491.1 &  & 6374.27 & 6403.29 & \textbf{363.9} &  & \textbf{6363.21} & \textbf{6384.12} & 487.1 \\
95 & 6223.54 & 6255.38 & 1332.4 &  & \textbf{6175.62} & \textbf{6232.75} & 476.9 &  & 6227.17 & 6233.96 & \textbf{241.9} \\ \hline
\multicolumn{1}{l}{\# of wins} & 2 & 3 & - &  & \textbf{16} & 7 & \textbf{29} &  & 13 & \textbf{20} & 1 \\
\multicolumn{1}{l}{APD} &  &  &  &  & \textbf{-0.54\%} & -0.32\% & \textbf{-167.21\%} &  & -0.51\% & \textbf{-0.50\%} & -147.24\% \\ \hline
\end{tabular}
\end{center}
\end{table}

\begin{table}[!ht] \scriptsize
\setlength{\tabcolsep}{5pt}
\begin{center}
\caption{Detailed comparison of MineReduce to other state-of-the-art algorithms -- Set 4 of \citet{duhameletal2010}}
\label{tab:SOTA4}
\begin{tabular}{crrrrrrrrrrr}
\hline
 & \multicolumn{3}{c}{HLS} & \multicolumn{1}{c}{} & \multicolumn{3}{c}{HILS-RVRP} & \multicolumn{1}{c}{} & \multicolumn{3}{c}{MineReduce} \\ \cline{2-4} \cline{6-8} \cline{10-12} 
I & \multicolumn{1}{c}{\begin{tabular}[c]{@{}c@{}}Best\\ Cost\end{tabular}} & \multicolumn{1}{c}{\begin{tabular}[c]{@{}c@{}}Avg.\\ Cost\end{tabular}} & \multicolumn{1}{c}{\begin{tabular}[c]{@{}c@{}}Avg.\\ Time\end{tabular}} & \multicolumn{1}{c}{} & \multicolumn{1}{c}{\begin{tabular}[c]{@{}c@{}}Best\\ Cost\end{tabular}} & \multicolumn{1}{c}{\begin{tabular}[c]{@{}c@{}}Avg.\\ Cost\end{tabular}} & \multicolumn{1}{c}{\begin{tabular}[c]{@{}c@{}}Avg.\\ Time\end{tabular}} & \multicolumn{1}{c}{} & \multicolumn{1}{c}{\begin{tabular}[c]{@{}c@{}}Best\\ Cost\end{tabular}} & \multicolumn{1}{c}{\begin{tabular}[c]{@{}c@{}}Avg.\\ Cost\end{tabular}} & \multicolumn{1}{c}{\begin{tabular}[c]{@{}c@{}}Avg.\\ Time\end{tabular}} \\ \hline
19 & 11760.02 & 11819.28 & 4849.1 &  & \textbf{11686.39} & 11745.69 & \textbf{236.4} &  & 11687.12 & \textbf{11725.38} & 676.4 \\
22 & 13240.93 & 13270.10 & 5921.0 &  & \textbf{13091.16} & \textbf{13134.19} & 658.7 &  & 13094.74 & 13145.70 & \textbf{587.6} \\
23 & 7769.33 & 7793.84 & 3658.6 &  & \textbf{7741.01} & \textbf{7782.68} & \textbf{329.8} &  & 7752.05 & 7783.74 & 487.5 \\
27 & \textbf{8417.62} & 8444.78 & 5349.5 &  & 8422.92 & 8442.97 & \textbf{320.1} &  & 8422.36 & \textbf{8441.02} & 598.1 \\
32 & 9378.30 & 9418.40 & 12004.5 &  & 9382.60 & 9436.70 & \textbf{440.3} &  & \textbf{9348.55} & \textbf{9404.36} & 1045.4 \\
38 & 11224.72 & 11271.90 & 4384.8 &  & 11194.68 & 11254.27 & \textbf{456.8} &  & \textbf{11192.74} & \textbf{11227.01} & 683.0 \\
46 & 24697.34 & 24832.07 & 8173.4 &  & 24566.23 & 24698.60 & \textbf{426.2} &  & \textbf{24404.42} & \textbf{24580.68} & 1345.2 \\
49 & 16282.44 & 16368.02 & 12796.0 &  & 16181.17 & 16322.51 & \textbf{559.7} &  & \textbf{16164.00} & \textbf{16257.90} & 1779.7 \\
58 & 23480.52 & \textbf{23543.71} & 6192.9 &  & \textbf{23370.42} & 23641.18 & \textbf{253.7} &  & 23396.28 & 23545.34 & 659.3 \\
62 & 23035.91 & 23094.21 & 6182.9 &  & 23010.35 & 23097.54 & \textbf{295.0} &  & \textbf{22903.99} & \textbf{23043.85} & 690.7 \\
65 & 13036.33 & 13102.95 & 5709.4 &  & 13043.54 & 13063.89 & \textbf{248.0} &  & \textbf{12975.38} & \textbf{13049.25} & 583.6 \\ \hline
\multicolumn{1}{l}{\# of wins} & 1 & 1 & - &  & 4 & 2 & \textbf{10} &  & \textbf{6} & \textbf{8} & 1 \\
\multicolumn{1}{l}{APD} &  &  &  &  & -0.36\% & -0.22\% & \textbf{-176.96\%} &  & \textbf{-0.53\%} & \textbf{-0.43\%} & -156.42\% \\ \hline
\end{tabular}
\end{center}
\end{table}

\clearpage

\bibliography{bibliography}

\end{document}